\documentclass[acmsmall]{acmart}

\AtBeginDocument{%
  }

\setcopyright{acmlicensed}
\acmJournal{PACMMOD}
\acmYear{2025} \acmVolume{3} \acmNumber{3 (SIGMOD)} \acmArticle{130} \acmMonth{6}\acmDOI{10.1145/3725394}
\newtheorem{definition}{Definition}

\usepackage{url}
\usepackage{cuted}
\usepackage{balance}
\usepackage[]{algpseudocode}
\usepackage{multirow}
\usepackage{multicol}
\usepackage{caption}
\usepackage{subcaption}
\usepackage{graphicx}
\usepackage{amsmath}
\usepackage{amsthm}
\usepackage{pifont}
\usepackage{algorithm}
\usepackage{tikz}
\usepackage{xcolor}

\begin{document}

\title{\textsf{Apt-Serve}: Adaptive Request Scheduling on Hybrid Cache for Scalable LLM Inference Serving}

\author{Shihong Gao}
\email{sgaoar@connect.ust.hk}
\affiliation{%
  \institution{The Hong Kong University of Science and Technology}
  \city{Hong Kong SAR}
  \country{China}
}

\author{Xin Zhang}
\affiliation{%
  \institution{The Hong Kong University of Science and Technology}
  \city{Hong Kong SAR}
  \country{China}
}
\email{sean.zhang@connect.ust.hk}

\author{Yanyan Shen}
\authornote{Yanyan Shen is the corresponding author.}
\affiliation{%
  \institution{Shanghai Jiao Tong University}
  \city{Shanghai}
  \country{China}
}
\email{shenyy@sjtu.edu.cn}

\author{Lei Chen}
\affiliation{%
  \institution{The Hong Kong University of Science and Technology (Guangzhou)}
  \city{Guangzhou}
  \country{China}
}
\affiliation{%
  \institution{The Hong Kong University of Science and Technology}
  \city{Hong Kong SAR}
  \country{China}
}
\email{leichen@cse.ust.hk}

\renewcommand{\shortauthors}{Shihong Gao, Xin Zhang, Yanyan Shen, \& Lei Chen}
\begin{abstract}
  Large language model (LLM) inference serving systems are essential to various LLM-based applications. As demand for LLM services continues to grow, scaling these systems to handle high request rates while meeting latency Service-Level Objectives (SLOs), referred to as effective throughput, becomes critical. However, existing systems often struggle to improve effective throughput, primarily due to a significant decline in Time To First Token (TTFT) SLO attainment. We identify two major causes of this bottleneck: (1) memory-intensive KV cache that limits batch size expansion under GPU memory constraints, and (2) rigid batch composition enforced by the default First-Come-First-Serve scheduling policy. In this paper, we introduce \textsf{Apt-Serve}, a scalable framework designed to enhance effective throughput in LLM inference serving. \textsf{Apt-Serve} features a new hybrid cache scheme that combines KV cache with a memory-efficient hidden cache for reusable input hidden state vectors, allowing large batch sizes and improving request concurrency. Based on the hybrid cache, \textsf{Apt-Serve} employs an adaptive runtime scheduling mechanism that dynamically optimizes batch composition. We formally define the adaptive scheduling optimization problem and propose an efficient algorithm with theoretical guarantees. Extensive evaluations on three real-world datasets and LLMs ranging from 13B to 66B parameters demonstrate that \textsf{Apt-Serve} achieves up to 8.8$\times$ improvement in effective throughput compared to the state-of-the-art inference serving systems.
\end{abstract}


\begin{CCSXML}
<ccs2012>
   <concept>
       <concept_id>10002951.10002952</concept_id>
       <concept_desc>Information systems~Data management systems</concept_desc>
       <concept_significance>500</concept_significance>
       </concept>
 </ccs2012>
\end{CCSXML}

\ccsdesc[500]{Information systems~Data management systems}

\keywords{request scheduling, cache management, inference serving}

\received{October 2024}
\received[revised]{January 2025}
\received[accepted]{February 2025}

\maketitle

\section{Introduction}
Large language models (LLMs)~\cite{achiam2023gpt, touvron2023llama, zhang2022opt,chowdhery2023palm, NEURIPS2020_1457c0d6, kaplan2020scaling, wei2022emergent} have emerged as a transformative force in artificial intelligence, demonstrating exceptional capabilities across a wide range of tasks including natural language understanding, question answering, and code generation.
This technological leap has catalyzed the development of LLM-based applications such as versatile chatbots~\cite{characterai, claude,achiam2023gpt}, advanced search engines~\cite{bing,team2023gemini,perplexity,komo}, and sophisticated programming assistants~\cite{dakhel2023github,replit,codewhisperer}. 

At the core of these applications are LLM inference serving systems, which generate highly contextual and coherent responses to varied user inputs.
Given an input request (i.e., a sequence of prompt tokens), the standard LLM inference process utilizes a Transformer-based decoder~\cite{transformer} and 
consists of two major phases. Initially, the \emph{prefill} phase processes the input prompt and generates the first response token. Subsequently, the \emph{decode} phase iteratively produces the next token based on the prompt and previously generated tokens until a termination token is encountered.
To ensure user satisfaction, LLM inference serving systems must adhere to Service-Level-Objectives (SLOs) that focus on two per-request latency metrics: i) \emph{Time To First Token} (TTFT), measuring the duration of the prefill phase; ii) \emph{Time Between Tokens} (TBT), quantifying the time between consecutive token generations during the decode phase.

With the growing demand for real-time LLM services, LLM inference serving systems have to continuously scale up to accommodate the increasing number of requests. 
This scaling challenge requires systems to enhance the
\textbf{{effective throughput}}~\cite{zhong2024distserve,agrawal2024taming,qin2024mooncake} which is defined as \emph{the highest sustainable online request rate that meets specified SLOs attainment criteria}, such as serving at least 70\% of requests within the target SLOs of (TTFT=1 second, 99th percentile TBT=1 second).
To achieve this, existing systems~\cite{kwon2023efficient,agrawal2024taming,zhong2024distserve,patel2024splitwise,qin2024mooncake} mainly adopt an iteration-level batching approach~\cite{yu2022orca} to process multiple requests on the GPU concurrently. 
Specifically, at the beginning of each iteration, a batch of requests is scheduled to the GPU for execution using the First-Come-First-Serve (FCFS) policy~\cite{kwon2023efficient,zhong2024distserve,agrawal2024taming}.
To reduce the computational cost of self-attention operations in LLMs, the systems implement KV cache~\cite{transformer} to store reusable key and value vectors for both the prompt and generated tokens in each Transformer layer. During execution, the GPU memory maintains the corresponding KV caches for all requests in the current batch. 
The iteration ends by generating one output token for every request within the batch. 
Some recent works~\cite{patel2024splitwise,agrawal2024taming,zhong2024distserve,oh2024exegpt,holmes2024deepspeed} focus on improving computation resource utilization by accounting for the different computational demands of the prefill and decode phases in LLM inference. They effectively mitigate interference between requests in different phases, reducing unnecessary SLO violations for both TTFT and TBT, thereby enhancing effective throughput. 

\begin{figure}
    \centering
    \includegraphics[width=0.7\columnwidth]{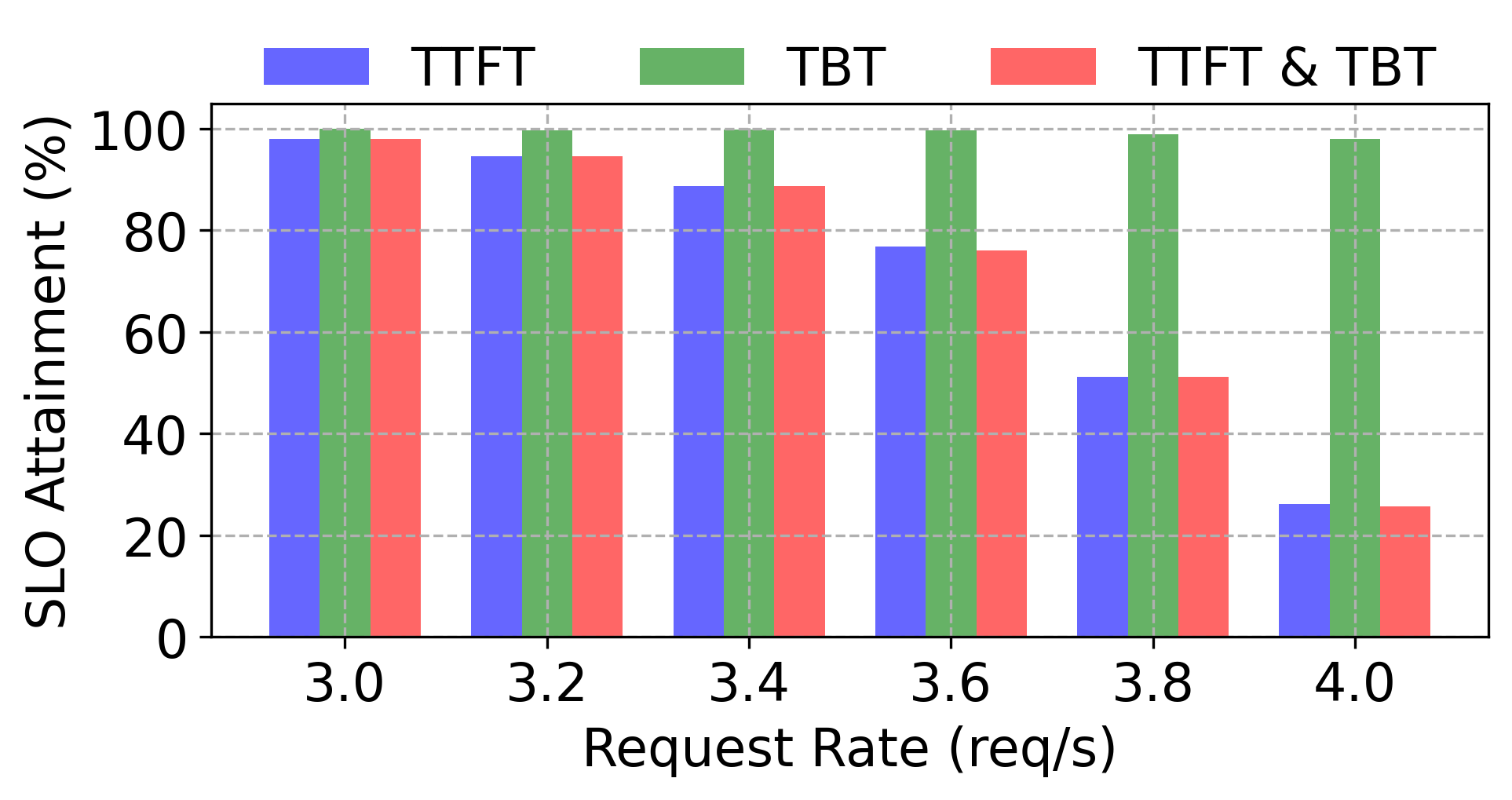}
    \caption{Serving sampled requests from the ShareGPT~\cite{sharegpt} dataset using the vLLM~\cite{kwon2023efficient} system with the OPT-13B model~\cite{zhang2022opt} on an NVIDIA A100 GPU. The X-axis represents the request rate (req/s), and the Y-axis shows the SLO attainment (\%) (TTFT: 1s, P99 TBT: 1s).}
    \label{fig:intro}
\end{figure}
Despite these advancements, we identify a performance wall that limits further effective throughput improvements of existing LLM inference serving systems. This wall stems from the system's inability to maintain TTFT SLO attainment as the request rate increases. As shown in Figure~\ref{fig:intro}, in a cutting-edge system~\cite{kwon2023efficient}, an increase in request rate significantly diminishes overall SLO attainment, i.e., most requests fail to meet both TTFT and TBT requirements. Notably, higher request rates cannot be sustained primarily due to a sharp decline in TTFT SLO attainment, while TBT SLO attainment remains largely unaffected. 
The question we would like to ask is: can we break through this performance wall and achieve a higher effective throughput of LLM inference service?

To answer the question, we identify two primary factors leading to the significant drop in TTFT SLO attainment in existing systems as the request rate increases. \textbf{{First}}, the extensive use of memory-intensive KV cache prevents further enlarged batch size under memory constraint. The size of the KV cache grows with the number of tokens processed per request. When the KV cache of ongoing requests nearly saturates the GPU memory, the system struggles to accommodate a larger batch size, leading to queuing delays for subsequent requests. As a result, frequent reaching of the batch size limit during serving can exacerbate queue delays, which causes more widespread TTFT SLO violations for incoming requests. Empirically, we observe that TTFT SLO attainment generally decreases as the system operates for longer periods at its maximum batch size capacity (Section~\ref{sc3.1}). \textbf{{Second}}, the default FCFS scheduling policy enforces rigid batch composition. When the system schedules a batch of requests for execution at the start of each inference iteration, the FCFS policy consistently prioritizes the earliest arriving requests, subject to the memory constraint of their cache storage. However, online requests often vary in sequence length, comprising different numbers of prompt and output tokens. This rigid FCFS policy limits the system’s flexibility to optimize batch composition, which could otherwise improve TTFT SLO attainment under the same request load. In practice, we observe that FCFS policy can even result in much worse TTFT SLO attainment compared to completely random scheduling under the same request rate for the same set of incoming requests (Section~\ref{sc3.2}).

In this paper, we introduce \textsf{Apt-Serve}, a new framework designed to enhance effective throughput in LLM inference serving. Building on the two previously discussed insights, \textsf{Apt-Serve} incorporates two innovative designs to tackle the bottleneck in existing LLM serving systems. \textbf{First}, to obtain a larger attainable batch size, \textsf{Apt-Serve} introduces a novel hybrid KV cache and hidden cache scheme. Specifically, both KV cache and hidden cache serve as reusable computation results during inference, while hidden cache stores input hidden vectors instead of intermediate key and value vectors for the self-attention in each Transformer layer. Hidden cache reduces cache memory consumption per request by half compared to the KV cache, at the expense of extra computational overhead when retrieving comprehensive past information. Therefore, when the system reaches its batch size limit under KV cache usage preventing subsequent requests from being served, \textsf{Apt-Serve} can reassign hidden cache usage in place of KV cache usage for some ongoing requests, and assign hidden cache for certain subsequent requests directly from the outset. Under the same memory consumption, this enables a larger batch size, reducing queuing delays for subsequent requests at a manageable cost of extra latency increase (TBT) for ongoing requests. \textbf{Second}, to optimize batch composition, \textsf{Apt-Serve} employs an adaptive runtime scheduling mechanism. As the \textit{full} processing time and memory usage of each request cannot be known in advance due to the non-deterministic final end token~\cite{kwon2023efficient}, \textsf{Apt-Serve} continuously monitors key runtime metrics for every request at hand in each inference iteration, such as their pending time and memory requirement so far. Based on such tracked runtime information, \textsf{Apt-Serve} formalizes the scheduling process in each inference iteration as an optimization problem, aiming to compose a batch of requests with appropriate cache type assignments to maximize the reduction in overall pending time, while respecting memory constraints for cache storage. By framing the per-iteration scheduling process this way, batch compositions are adaptively adjusted to account for changing runtime conditions across iterations. As the formulated optimization problem is NP-hard, \textsf{Apt-Serve} adopts an efficient greedy-based solution, for which we establish a theoretical approximation ratio of 2.

We implement \textsf{Apt-Serve} atop the state-of-the-art LLM serving system vLLM~\cite{kwon2023efficient}, seamlessly incorporating all previously mentioned designs. We add additional supports including runtime information tracking and quantification to facilitate adaptive scheduling design, and implement a tailored block-wise storage scheme with customized CUDA kernels to streamline the hybrid cache usage.  

In summary, this paper makes the following major contributions:
\begin{itemize}
 \item We pinpoint two key factors that prevents the existing system from delivering higher effective throughput: (1) the exhaustive use of memory intensive-KV cache limits the batch size, and (2) the rigid FCFS scheduling policy causes suboptimal batch composition.
 \item We introduce a new LLM inference serving framework named \textsf{Apt-Serve} that optimizes effective throughput. It employs a new hybrid cache scheme to enlarge batch size, and an innovative adaptive scheduling mechanism to optimize the batch composition based on runtime information.
 \item We formulate the per-iteration scheduling process in \textsf{Apt-Serve} as an optimization problem, allowing batch compositions to be adaptively refined based on changing runtime conditions. We show the problem is NP-Hard and propose an efficient greedy-based solution with a theoretical guarantee. 
 \item We conduct extensive experimental evaluations to validate the effectiveness of \textsf{Apt-Serve} using three real-world datasets that correspond to diverse application scenarios, as well as across LLMs varying in size from 13B to 66B parameters. Compared to state-of-the-art systems, \textsf{Apt-Serve} enhances the effective throughput by up to 8.8$\times$.
\end{itemize}

\section{Preliminaries}\label{bg}
In this section, we first introduce the transformer-based large language models. We further illustrate the necessary preliminaries for the online LLM inference serving. Table~\ref{tab:notations} summarizes the key notations used throughout this paper.
\subsection{Transformer-based Large Language Models} \label{trans_model}
Prevalent large language models (LLMs) such as GPT~\cite{achiam2023gpt}, OPT~\cite{zhang2022opt} and LLaMA~\cite{touvron2023llama} are decoder-only Transformer~\cite{transformer} models designed for the next-token prediction task. Generally, these LLMs are composed of an input embedding layer, followed by a series of Transformer layers, and finally end with an output projection layer. The input embedding layer converts a sequence of tokens $T=(t_1, t_2, ..., t_n)$ into a sequence of \emph{initial input vectors} $\mathbf{X}=(\mathbf{x}_1, \mathbf{x}_2, ..., \mathbf{x}_n)$. Then, the stack of Transformer layers processes the initial input vectors $\mathbf{X}$ to generate the output vectors $\mathbf{O}=(\mathbf{o}_1, \mathbf{o}_2, ..., \mathbf{o}_n)$. Finally, the output projection layer projects the last output vector $\mathbf{o}_n$ into logits for predicting the next token $t_{n+1}$. As an essential component in LLMs, each Transformer layer consists of two key modules: the self-attention module and the feed-forward network module. 
Let $\mathbf{X}^{\ell}=(\mathbf{x}^{\ell}_1, \mathbf{x}^{\ell}_2, \dots, \mathbf{x}^{\ell}_n) \in \mathbb{R}^{n\times d}$ be the sequence of \textbf{input hidden state vectors} at the $\ell$-th Transformer layer.

\noindent \textbf{Self-attention Module.} This module aims to generate contextualized vector representations by modeling complex correlations among the vectors. It first performs three linear transformations on each input hidden state vector $\mathbf{x}_i^\ell$ in $\mathbf{X}^{\ell}$ to obtain the \emph{query, key} and \emph{value} vectors, using the following equations.
\begin{align} 
\mathbf{q}^{\ell}_{i}=\mathbf{W}^{\ell}_{q}{\mathbf x}^{\ell}_{i}, \quad
\mathbf{k}^{\ell}_{i}=\mathbf{W}^{\ell}_{k}{\mathbf x}^{\ell}_{i}, \quad
\mathbf{v}^{\ell}_{i}=\mathbf{W}^{\ell}_{v}\mathbf{x}^{\ell}_{i}. \label{qkv_proj}
\end{align}
It then computes the scaled dot-product of each query vector $\mathbf{q}_i^\ell$ with all the preceding key vectors $\{\mathbf{k}_j^\ell\}_{j=1}^i$, and obtain the normalized attention scores $\{a_{ij}^\ell\}_{j=1}^i$ as follows.
\begin{align}
a^{\ell}_{ij} &= \frac{\mathrm{exp}\left({\mathbf{q}^{\ell}_{i}}^{\top} \mathbf{k}^{\ell}_{j}/\mathrm{\sqrt{d}}\right)}{\sum_{m=1}^{i} \mathrm{exp}\left({\mathbf{q}^{\ell}_{i}}^{\top} \mathbf{k}^{\ell}_{m}/\mathrm{\sqrt{d}}\right)}. \label{q_k}
\end{align}
Finally, the output vector $\mathbf{o}_i^\ell$ is computed by performing a weighted sum of the value vectors $\{\mathbf{v}_j^\ell\}_{j=1}^i$, which is defined as:
\begin{align}
 \mathbf{o}^{\ell}_{i} &= \mathbf{W}^{\ell}_{o}\sum_{j=1}^{i} a^{\ell}_{ij}\mathbf{v}^{\ell}_{i}.\label{a_v}
\end{align}

In this way, at each token position $i$ ($i=1,\dots, n$), self-attention computation is performed by attending to all of the preceding tokens and itself, to generate the corresponding output vector representation. This leads to a computational complexity of $\mathcal{O}(n^2)$, which scales quadratically with the number of input tokens.

\noindent \textbf{Feed-Forward Network (FFN) Module.} After the self-attention module, the feed-forward network uses an MLP layer followed by a linear transformation to obtain the final output vectors $\{{\mathbf{h}^{\ell}_{i}}\}_{i=1}^n$ of the $\ell$-th Transformer layer:
\begin{align}
\mathbf{z}^{\ell}_{i} = \mathrm{\sigma}\left(\mathbf{W}^{\ell}_{z} \mathbf{o}^{\ell}_i\right), \quad
\mathbf{h}^{\ell}_{i} = \mathbf{W}^{\ell}_{h} \mathbf{z}^{\ell}_{i}, \label{ffn_proj}
\end{align}
where $\mathrm{\sigma}$ denotes an activation function such as ReLU~\cite{agarap2018deep}.
Typically, $\mathbf{W}^{\ell}_{z}$ expands the attention output $\mathbf{o}^{\ell}_{i}$ from the dimension $d$ to a higher dimension $d^{'}$ (i.e., $d^{'}>d$), and $\mathbf{W}^{\ell}_{h}$ projects the intermediate vector $\mathbf{z}^{\ell}_{i}$ back to the dimension $d$. 

\begin{table}[t]
    \small
    \centering
    \caption{Summary of key notations.}
    \scalebox{0.9}{
    \begin{tabular}{c|c}
        \toprule
        $\mathbf{x}^{\ell}_i$ & input hidden state vector for token $i$ in the $\ell$-th Transformer layer\\
        \midrule
        $\mathbf{q}^{\ell}_i$, $\mathbf{k}^{\ell}_i$, $\mathbf{v}^{\ell}_i$ & query, key, value vectors for token $i$ in the $\ell$-th Transformer layer\\
        \midrule
        $\mathbf{h}^{\ell}_i$ & output vector for token $i$ in the $\ell$-th Transformer layer\\
        \midrule
        $\mathrm{W}^e$ & waiting queue of requests at iteration $e$\\
        \midrule
        $\mathrm{R}^e$ & running queue of requests at iteration $e$\\
        \midrule
        $\mathrm{U}^e$ & set of candidate requests to be scheduled at iteration $e$\\
        \midrule
        $\mathrm{M}^e$ & memory constraint for cache storage at iteration $e$\\
        \midrule
        $\mathrm{m}^{e}_{i}$ & maximum memory requirement by request $i$ at iteration $e$\\
        \midrule
        $\mathrm{p}^{e}_{i}$ & pending time of request $i$ at iteration $e$ \\
        \midrule
        $\mathrm{g}^{e}_{i}$ & the value of scheduling request $i$ at iteration $e$ \\
        \midrule
        $\alpha^{e}_i$ & binary variable for scheduling request $i$ at iteration $e$\\
        \midrule
        $\beta^{e}_i$ & binary variable for hidden cache usage of request $i$ at iteration $e$\\
        \bottomrule 
    \end{tabular}}
    \label{tab:notations}
\end{table}
\subsection{Generative LLM Inference Serving}\label{serving_pre}
\textbf{Auto-regressive Inference.} Nowadays LLMs~\cite{achiam2023gpt,zhang2022opt,touvron2023llama} employ an auto-regressive inference fashion. Given a request $r$ consisting of a sequence of tokens $(t_1, t_2, ..., t_n)$ as the prompt, an LLM generates the output tokens $(t_{n+1}, t_{n+2}, ..., t_{n+T})$ sequentially until an “end-of-sequence” (EOS) termination token is produced.
This process is divided into two phases. The \textit{prefill} phase involves a single iteration where LLM processes the entire prompt $(t_1, t_2, ..., t_n)$ to generate the first output token $t_{n+1}$. The \textit{decode} phase performs multiple iterations where LLM repeatedly takes the prompt tokens plus the previously generated output tokens as input and generates the next output token.  
Essentially, the model generates subsequent tokens based on the continuously expanding context.

\noindent \textbf{KV Cache Technique.} The KV cache technique optimizes inference at a target token position $i$ by storing the key vectors $\mathbf{K}^{\ell}_{i} = (\mathbf{k}^{\ell}_{1}, \dots, \mathbf{k}^{\ell}_{i-1})$ and value vectors $\mathbf{V}^{\ell}_{i} = (\mathbf{v}^{\ell}_{1}, \dots, \mathbf{v}^{\ell}_{i-1})$ from previous token positions in GPU memory for each Transformer layer $\ell = 1, \dots, L$. Without the KV cache, every token position $j$ ($j = 1, 2, \dots, i$) in a given layer $\ell$ must go through all computations from Eq.\ref{qkv_proj} to Eq.\ref{ffn_proj} for each decode iteration across all layers. This is due to the cascading nature of Transformer layers: the computation of the next layer $\ell+1$ still requires the input hidden vectors $\mathbf{x}^{\ell+1}_{j}$ for all token positions ($j = 1, 2, \dots, i$), which are derived from the output $\mathbf{h}^{\ell}_{j}$ of the current layer $\ell$. By caching $\mathbf{K}^{\ell}_{i}$ and $\mathbf{V}^{\ell}_{i}$ at each layer, only the current token position $i$ needs to undergo the computation from Eq.\ref{qkv_proj} to Eq.\ref{ffn_proj}. This reduces the complexity of self-attention (Eq.\ref{q_k}-\ref{a_v}) from $\mathcal{O}(n^2)$ to $\mathcal{O}(n)$, and the complexity of other operations (Eq.\ref{qkv_proj} and Eq.\ref{ffn_proj}) from $\mathcal{O}(n)$ to $\mathcal{O}(1)$ per layer during each decoding iteration.

\noindent \textbf{Block-wise KV Cache Storage.} 
Existing LLM inference serving systems~\cite{kwon2023efficient,zhong2024distserve,agrawal2024taming,Llumnix, oh2024exegpt, patel2024splitwise} adopt a block-wise storage approach~\cite{kwon2023efficient} to optimize the GPU memory utilization for KV cache. The earlier approach~\cite{yu2022orca} pre-allocates a contiguous memory space for each request's KV cache storage up to the maximum sequence length by the model. Such pre-allocation allows for fast KV cache retrieval during inference. However, since the actual output length of each request is unpredictable and typically shorter than the maximum, this method can lead to internal memory fragmentation~\cite{kwon2023efficient}, which limits the batch size. In contrast, block-wise KV cache storage divides the total available GPU memory into fixed-size blocks. As the output length increases, cache blocks are allocated on demand. In this way, the cache blocks for the same request may scatter across different physical locations.
Such a strategy improves memory efficiency by reducing internal fragmentation, thus enabling larger batch sizes for concurrent request processing.
However, it is crucial to notice that KV cache space still grows linearly to the sequence length. Its memory-intensive nature still imposes a burden for GPU memory consumption and inevitably constrains batch size. 

\noindent \textbf{Iteration-level Batching in LLM Inference Serving.} 
LLM inference serving has been a critical workload in modern data centers~\cite{Llumnix}. It is thus required to handle dynamically arriving requests with varying sequence lengths. 
To achieve this, existing systems perform iteration-level batching~\cite{yu2022orca} that allows a new request to join the ongoing batch and a finished request to leave the batch at each inference iteration.
Specifically, at the beginning of each inference iteration, the serving system assesses available GPU memory for KV cache storage of the requests at hand. It then decides whether to admit some new requests for a prefill iteration or continue with a decode iteration for active requests.
Each request in the batch has its associated KV cache stored in GPU memory, which expands incrementally as a new output token is generated.
Our proposed \textsf{Apt-Serve} framework adopts the standard iteration-level batching, but introduces an innovative adaptive scheduling mechanism that incorporates a new hybrid cache scheme.

\begin{figure}[t]
    \centering
    \begin{subfigure}[b]{0.35\columnwidth}
        \centering
        \includegraphics[width=\textwidth]{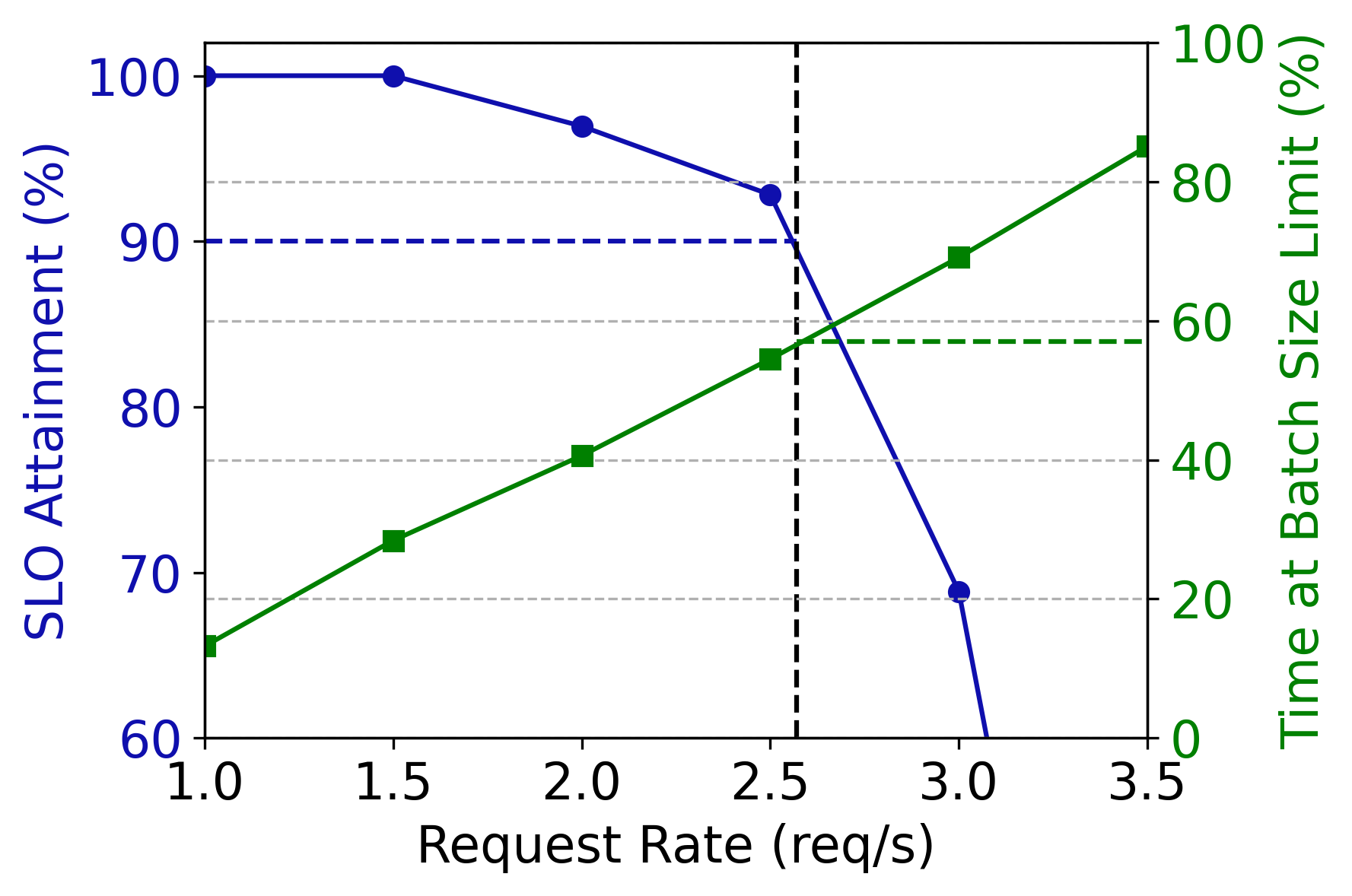}
        \caption{}\label{kv_cache_result}
    \end{subfigure}
    \begin{subfigure}[b]{0.35\columnwidth}
        \centering
        \includegraphics[width=\textwidth]{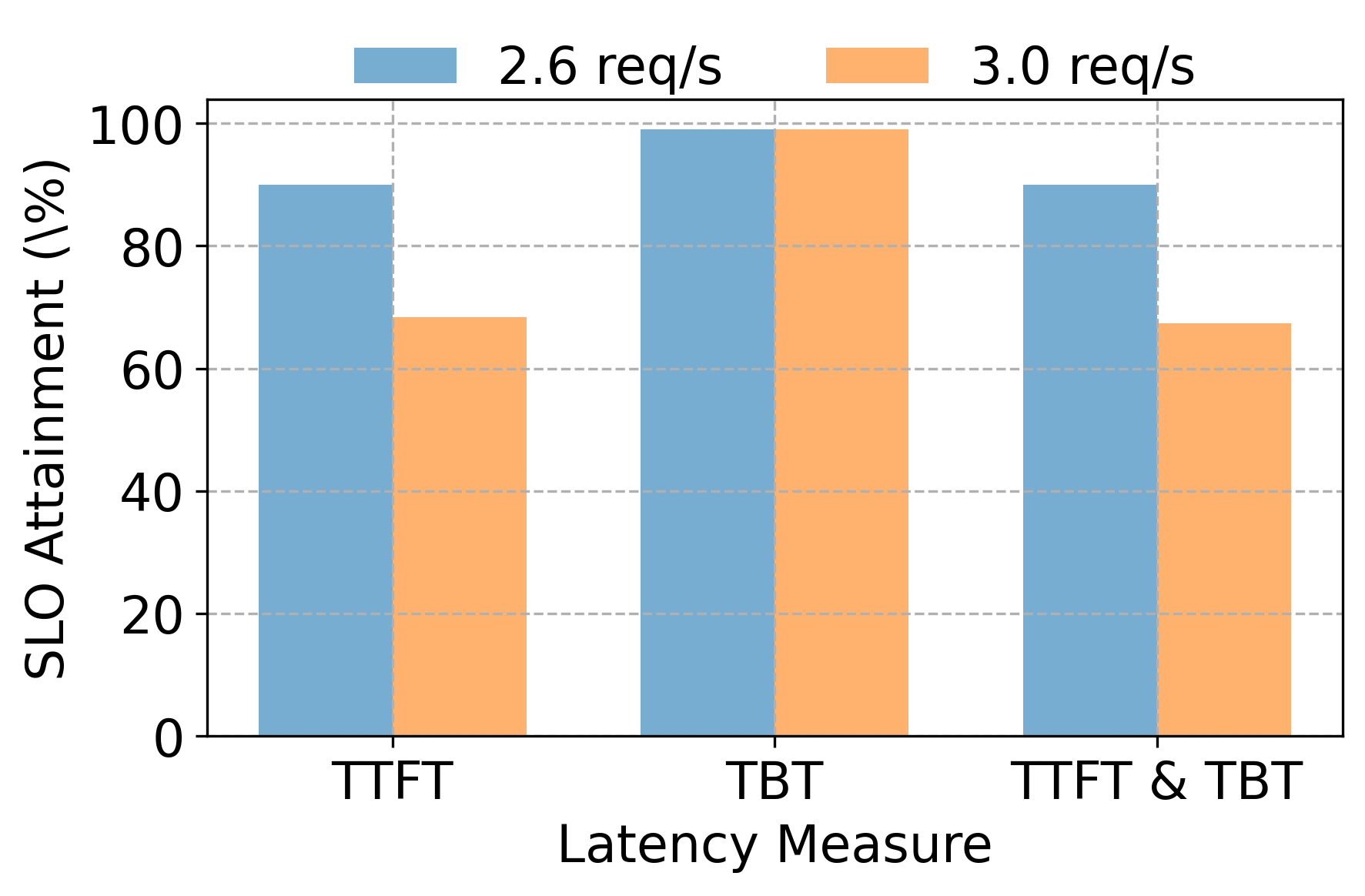}
        \caption{}\label{pre_part2}
    \end{subfigure}
    \caption{(a) SLO attainment rate (\%) and time ratio at batch size limit (\%) under varying request rates. The left Y-axis is the percentage of served requests adhering to the SLOs, while the right Y-axis is the percentage of serving time the system operates at the batch size limit. The X-axis is the varied request rates. (b) A comparison of specific SLO attainments at request rates of 2.0/reqs and 3.0 req/s.}
\end{figure}

\section{Bottlenecks and Opportunities} \label{sc3}
We focus on improving the effective throughput of LLM inference services, which currently experience performance bottlenecks due to the rapid decline in TTFT SLO attainment as the request rate increases. 
Through a close look at the practical inference process, we attribute the TTFT SLO violation to two critical factors: (1) the heavy reliance on memory-intensive KV cache which restricts batch size, and (2) the inflexibility of the FCFS scheduling policy which often results in suboptimal batch compositions.
In this section, we provide an in-depth empirical analysis to disclose the performance bottlenecks caused by the two factors and identify the potential opportunities to mitigate their impact on the effective throughput.

\begin{figure}[t]
    \centering
    \begin{subfigure}[b]{.49\columnwidth}
        \centering
        \includegraphics[width=1.0\textwidth]{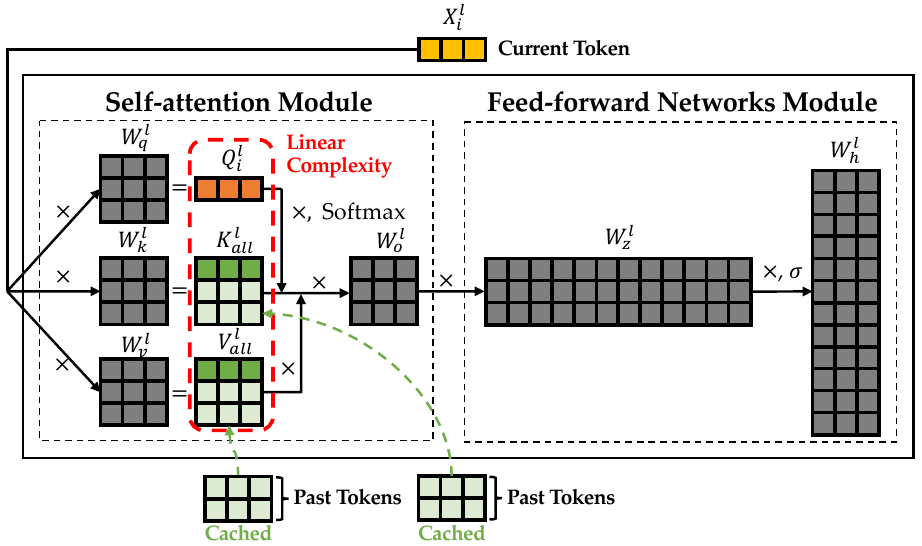}
        \caption{Decoding with KV cache in a Transformer layer}\label{kv_cache}
    \end{subfigure}
    \begin{subfigure}[b]{.49\columnwidth}
        \centering
        \includegraphics[width=1.0\textwidth]{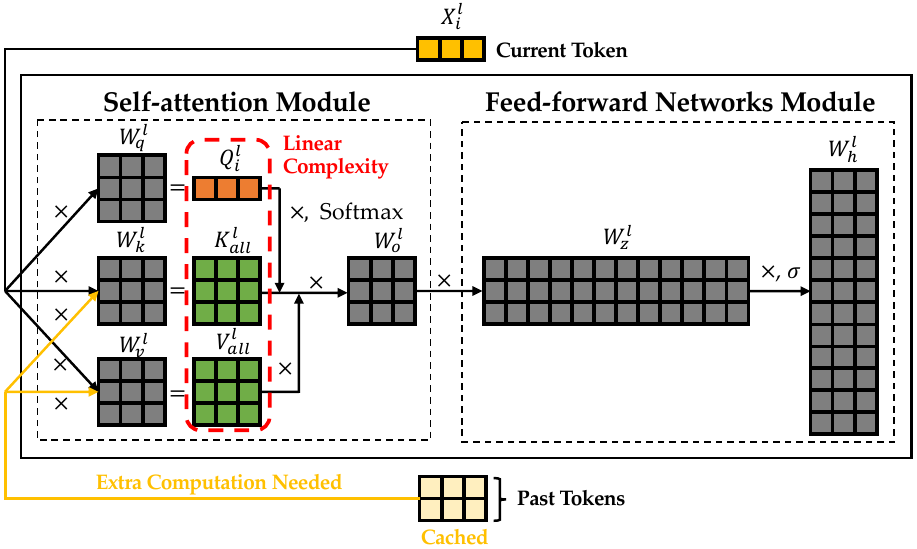}
        \caption{Decoding with hidden cache in a Transformer layer}\label{hidden_cache}
    \end{subfigure}
    \caption{The illustrations of the computations performed during the decode phase for a request (a) with KV cache and (b) with hidden cache. Assume the original input hidden state vectors for the target request consist of a length of 3 (1 current token + 2 past tokens), and each vector has a dimension of 3.
    }
    \label{fig:kv_vs_hidden}
\end{figure}

\begin{figure}[t]
	\centering
	\begin{subfigure}[b]{0.32\columnwidth}
		\centering
		\includegraphics[width=\textwidth]{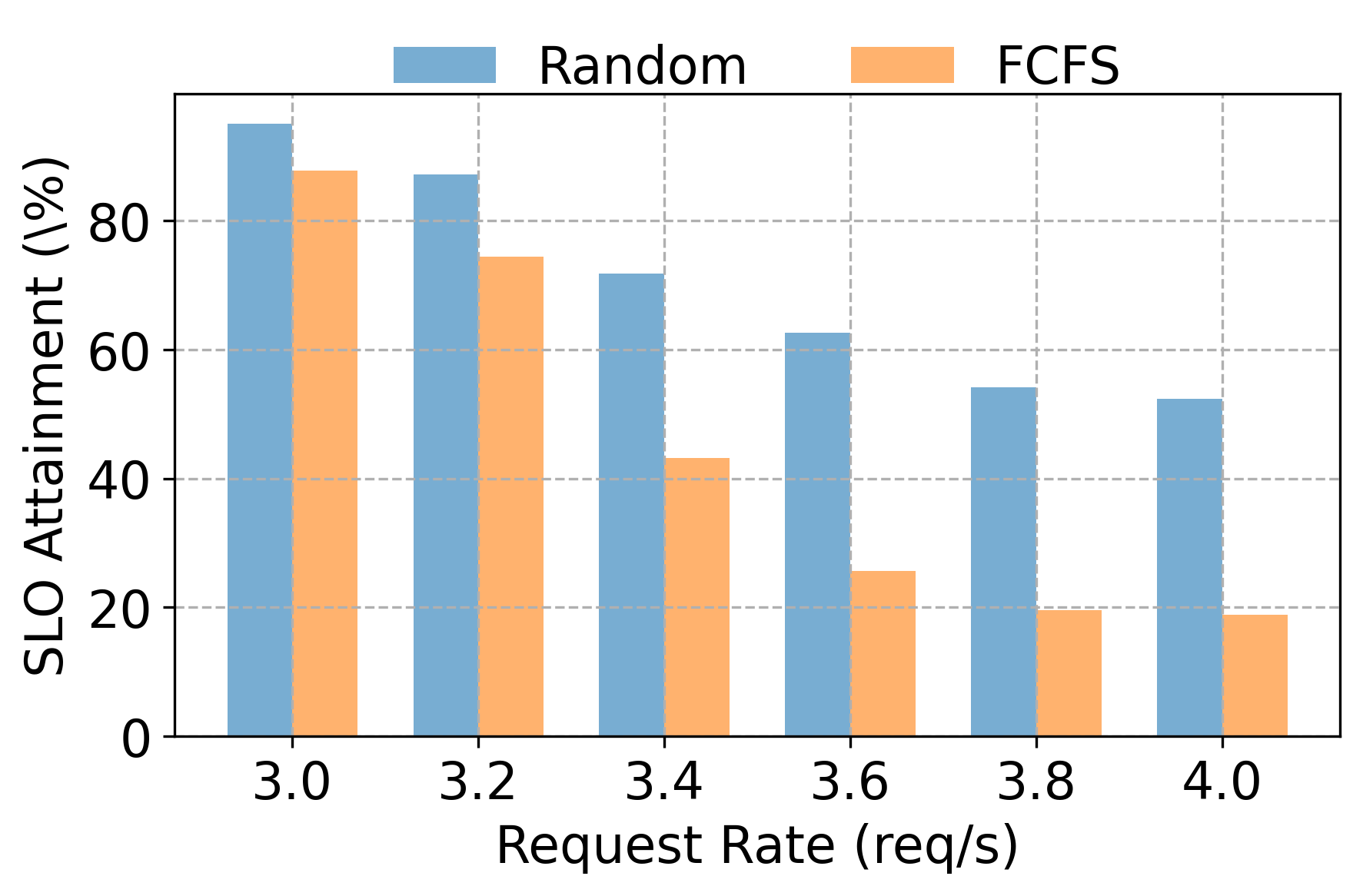}
		\caption{SLO attainment comparison}\label{pre2_comp}
	\end{subfigure}
	\begin{subfigure}[b]{0.32\columnwidth}
		\centering
		\includegraphics[width=\textwidth]{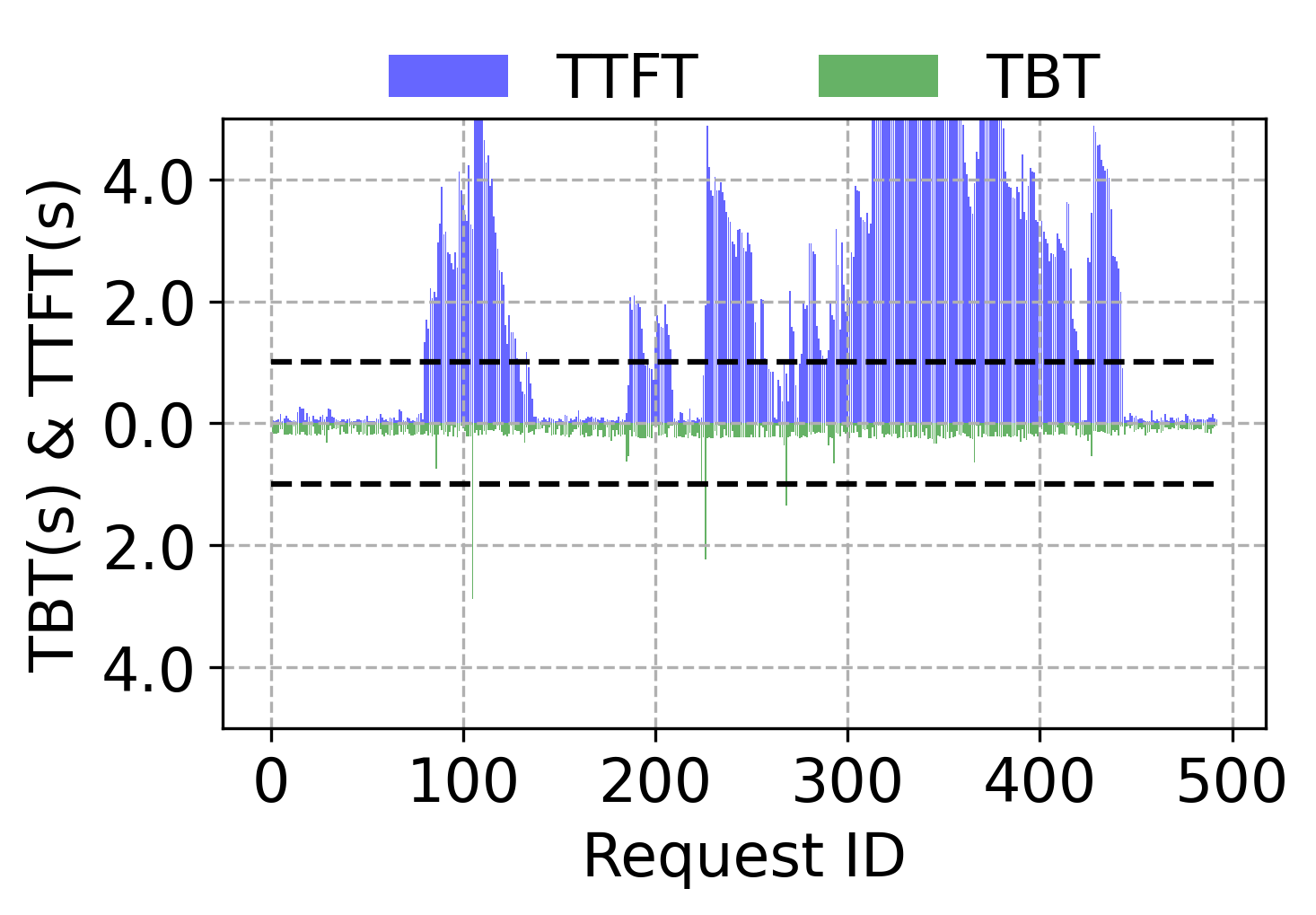}
		\caption{FCFS scheduling}\label{slo_fcfs}
	\end{subfigure}
	\begin{subfigure}[b]{0.32\columnwidth}
		\centering
		\includegraphics[width=\textwidth]{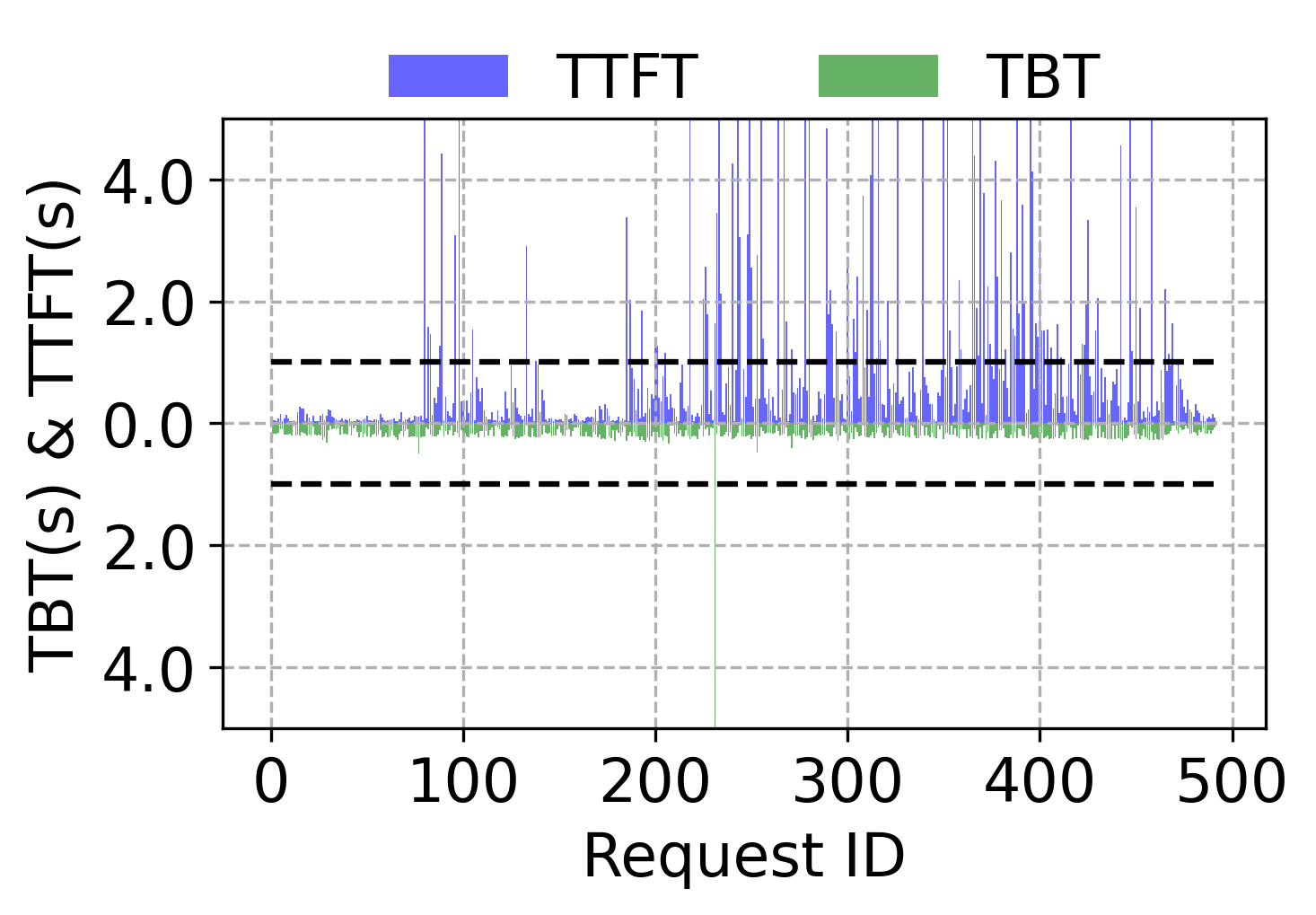}
		\caption{Random scheduling}\label{slo_random}
	\end{subfigure}
	\caption{(a) SLO attainment (\%) comparison between FCFS and random scheduling policies. The X-axis represents the request rate, and the Y-axis shows the percentage of requests meeting SLOs. (b) Distribution of per-request SLO attainment using FCFS scheduling. (c) Distribution of per-request SLO attainment using random scheduling. In (b) and (c), the X-axis displays request IDs sorted by arrival time, the upper Y-axis is the TTFT latency, and the lower Y-axis is the P99 TBT latency for served requests.}
\end{figure}
\subsection{Memory-Intensive KV Cache Usage}\label{sc3.1}
It has been recognized that KV caches tend to consume high GPU memory space~\cite{kwon2023efficient}, especially when dealing with long-latency requests, i.e., those with extensive prompts or lengthy output sequences.
The memory-intensive nature of KV cache causes limited batch size, thereby reducing the parallelism in request processing and delaying the serving of subsequent requests. These deplayed requests are more likely to voilate the specified TTFT SLO constraint and hurts the effective throughput of the system. 

To assess the effect of the memory-intensive KV cache usage on the inference performance, we simulate a workload of 500 requests, randomly sampled from the ShareGPT dataset~\cite{sharegpt}, with request arrivals following a Poisson distribution. 
The requests are served using the OPT-13B model~\cite{zhang2022opt} on an NVIDIA A100 GPU with 40GB memory. We employ a state-of-the-art inference serving system~\cite{kwon2023efficient} with block-wise KV cache management enabled. 
We vary the request rates and set the latency SLOs to 1 second for both TTFT and P99 TBT\footnote{P99 TBT refers to the 99th percentile of TBT latency for each individual request, where the generation of one output token in the decode phrase contributes to a TBT latency value~\cite{agrawal2024taming,qin2024mooncake}.}. We record the TTFT SLO attainment (\%) and measure the proportion of total serving time
during which the batch size is at its maximum capacity. This maximum is determined by the GPU memory constraint at runtime, beyond which the batch size cannot be increased further.

Figure~\ref{kv_cache_result} shows that
with a 90\% SLO attainment threshold, the effective throughput is around 2.6 requests per second. For 60\% of the overall serving time, i.e., the time to finish all the incoming requests' generation, the batch size cannot be increased to accommodate more requests regarding the available GPU memory space. 
When the request rate reaches 3 requests per second, the system hits the batch size limit for over 80\% of the serving time (due to insufficient space for storing additional requests' cache), causing the SLO attainment to drop sharply to approximately 70\%.
Figure~\ref{pre_part2} compares the SLO attainments at two request rates, 2.6 req/s and 3 req/s, indicating the SLO violation is mainly due to the decline in TTFT SLO attainment. These findings suggest that hitting the batch size limit frequently hampers concurrent request processing, resulting in many requests being delayed and failing to meet the TTFT requirements.

\textbf{Opportunity I: Utilizing Hybrid Cache.}
We observe that the hidden state vectors $\mathbf{X}_{i}^{\ell} = (\mathbf{x}^{\ell}_1, \mathbf{x}^{\ell}_2, \dots, \mathbf{x}^{\ell}_{i-1})$, which serve as the input at any given $\ell$-th Transformer layer ($\ell=1,2,...,L$) are also reusable intermediate results to address quadratic self-attention complexity, as we can temporarily transform the cached input hidden vectors $\mathbf{X}_{i}^{\ell}$ to required key and value vectors $\mathbf{K}_{i}^{\ell}$ and $\mathbf{V}_{i}^{\ell}$ by the self-attention operation on the fly (illustrated by the yellow lines in Figure~\ref{hidden_cache}). Such input hidden state vectors demands half the storage space compared to the key and value vectors (Figure~\ref{kv_cache}).
This inspires us to develop a hidden cache (Figure~\ref{hidden_cache}) for storing reusable input hidden state vectors as an alternative to the KV cache for certain requests. When the system hits its batch size limit due to the extensive use of KV caches, it becomes advantageous to 1) convert the KV caches of some ongoing requests to hidden caches, and 2) assign hidden caches to new requests, allowing them to begin their prefill phase without delay. As a benefit, the system is allowed to expand its batch size to promote the overall effective throughput. 
However, hidden cache usage causes $\mathcal{O}(n)$ complexity for key and value linear projection in each Transformer layer (Eq.\ref{qkv_proj}) required for restoring the key and value vectors, as opposed to $\mathcal{O}(1)$ by the usage of direct KV cache. This extra step may decrease batch processing speed. To this end, we introduce an effective hybrid cache scheme to balance batch size and processing speed. We dynamically determine the optimal timing and cache type allocation for each request to maximize the effective throughput.

\subsection{Rigid FCFS Scheduling Policy}\label{sc3.2}
Existing LLM inference serving systems~\cite{kwon2023efficient,agrawal2024taming,zhong2024distserve,patel2024splitwise,qin2024mooncake} typically employ the First-Come-First-Serve (FCFS) scheduling policy to form a batch for each inference iteration. This method prioritizes requests based solely on their arrival time, overlooking the variability in prompt lengths or expected output sizes across different requests. Such rigid scheduling may lead to suboptimal batch compositions, undermining the overall inference performance of the system.
To evaluate how different scheduling policies affect system performance, we substitute the original FCFS scheduling with a random scheduling method in the cutting-edge system~\cite{kwon2023efficient}. We use the same simulation setup as described in Section~\ref{sc3.1}, and compare the effective throughput of the system under both scheduling methods. To ensure a fair comparison, we maintain identical request arrival sequences across various request rates. 

The comparison results are shown in Figure~\ref{pre2_comp}. Notably, random scheduling consistently achieves higher SLO attainment than FCFS scheduling across all request rates. This suggests that FCFS scheduling leads to suboptimal batch compositions. To delve deeper into this observation, we examine the distribution of SLO attainment on a per-request basis under two scheduling policies with a request rate of 3.4 req/s. Figures~\ref{slo_fcfs} and~\ref{slo_random} visualize the distributions for FCFS and the random scheduling policies, respectively. It is evident that random scheduling leads to fewer TTFT SLO violations. Moreover, we notice that under FCFS scheduling, TTFT SLO violations tend to occur in clusters of consecutive requests, whereas these violations are more evenly distributed under random scheduling. 

\textbf{Opportunity II: Runtime Dynamic Scheduling.} 
The potential benefits of random scheduling inspire us to develop a more sophisticated scheduling policy, leading to better batch compositions that reduce TTFT SLO violations, e.g., it may dynamically initiate the prefill phase for certain new requests while earlier ones are still in their decode phase, recalling that a higher request rate generally does not affect the TBT SLO attainment (see Figure~\ref{fig:intro} and Figure~\ref{pre_part2}). 
However, determining the ideal optimal batch composition through runtime dynamic scheduling is a non-trivial task. The difficulties arise from 1) the dynamic nature of request arrivals in an online serving environment, and 2) the uncertainty of each request's duration, due to the unpredictable output lengths in auto-regressive LLM inference. In this paper, we propose a novel scheduling mechanism that adapts to the system's runtime conditions and request characteristics, leading to more flexible batch compositions.

\begin{figure}[t]
    \centering
    \includegraphics[width=0.6\columnwidth]{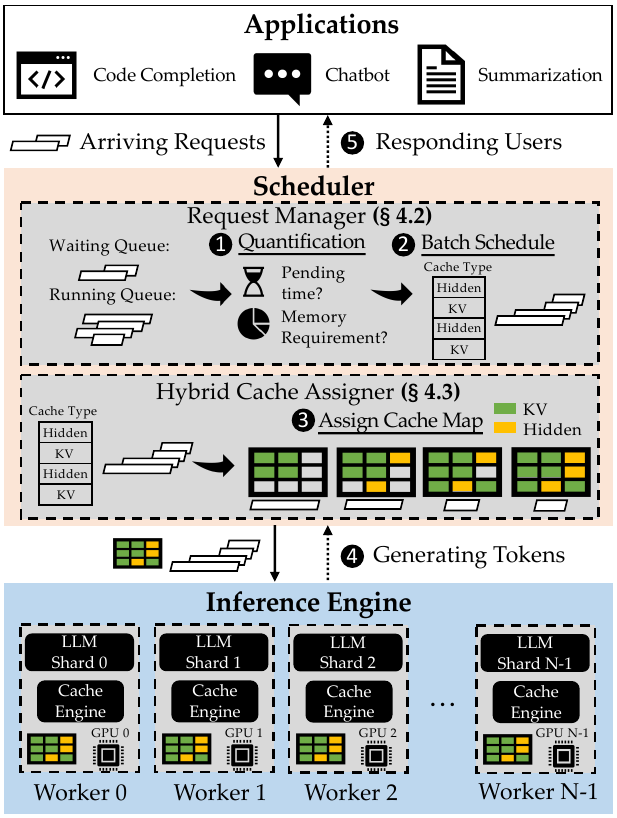}
    \caption{An overview of \textsf{Apt-Serve}.}
    \label{fig:overview}
\end{figure}
\section{Apt-Serve}
In this section, we first present an overview of our \textsf{Apt-Serve} framework. We then elaborate on the details of the request manager and the hybrid cache assigner with the cache engine. 

\subsection{Framework Overview}\label{overview}
Figure~\ref{fig:overview} presents an overview of \textsf{Apt-Serve}. Its high-level architecture aligns with standard practices in existing serving systems~\cite{kwon2023efficient, agrawal2024taming, zhong2024distserve, oh2024exegpt, Llumnix}, consisting of an iteration-level batch scheduler and an inference engine that work in close coordination to stream output tokens one by one to each user. The request manager is responsible for determining the batch schedule at the start of each inference iteration, while the inference engine assigns GPU workers to efficiently process the scheduled batch of requests. In each iteration, the workers perform a pass over the model parameters to generate one output token in parallel for each request within the batch. Within the scheduler and the inference engine, \textsf{Apt-Serve} further incorporates more sophisticated lower-level designs. The request manager inside the scheduler considers crucial factors such as pending time and memory requirements to optimize batch composition for each inference iteration. The tailored hybrid cache manager within the scheduler and the customized cache engine equipped in the inference engine help to streamline the novel usage of hybrid KV cache and hidden cache during inference.

\subsection{Request Manager}\label{request_manager}
The request manager maintains a waiting queue $\mathrm{W}^e$ and a running queue $\mathrm{R}^e$ for each target inference iteration $e$, where $e = 1, \dots, \infty$. The waiting queue $\mathrm{W}^e$ includes requests that are currently on hold due to their cache being unavailable in GPU memory. Specifically, this comprises requests that have arrived at the system but have not yet entered the prefill phase (with no output tokens received), and those that are in the decoding phase but were preempted in earlier iterations\footnote{These preempted requests can be resumed via a prefill iteration, with their initial prompt tokens and previously generated output tokens as the new input.}. Conversely, the running queue $\mathrm{R}^e$ contains requests in the decode phase, with their associated cache currently in the GPU memory. Drawing inspirations from the cost/value-based scheduling widely adopted in data management community~\cite{wang2016database,li2021ai,gupta2009fair,chi2013distribution, wagner2021self, cao2023transaction, mao2023morphstream, cheng2024towards, bouganim2000dynamic}, the request manager develops a quantification model along with an adaptive runtime scheduling mechanism to optimize batch composition during serving.

\textbf{Workflow.} In each inference iteration, it utilizes the tracked runtime information of the relevant requests as input, which is then processed by the quantification model to calculate the value of each candidate schedule for the current iteration. Ultimately, through the adaptive runtime scheduling mechanism, the request manager generates the final batch schedule by selecting from all candidate schedules based on their associated quantified values.

\textbf{Runtime Information Tracking.} As the arrivals and lifetime of online requests are not known as a priori (Section~\ref{sc3.2}), the request manager continuously tracks the available runtime information for every request at hand in each inference iteration. Specifically, at a target iteration $e$, for a candidate request $i$ from either the waiting queue $\mathrm{W}^{e}$ or running queue $\mathrm{R}^{e}$, the request manager records its maximum memory requirement $\mathrm{m}^{e}_{i}$ (i.e., size of KV cache rather than hidden cache) and pending time $\mathrm{p}^{e}_{i}$ so far. For the pending time $p^{e}_{i}$, if the request $i$ has not entered the prefill phase, its pending time $\mathrm{p}^{e}_{i}$ is calculated as the current time minus the time it arrives. Otherwise, its pending time $p^{e}_{i}$ is calculated as the current time minus the last time it has received an output token. 

\textbf{Quantification Model.} The quantification model within the request manager leverages the tracked runtime information for any given request $i,\forall i \in \mathrm{W}^e \cup \mathrm{R}^e$, in any iteration $e, \forall e=1,2,...,\infty$, to explicitly quantify the value $\mathrm{g}^{e}_i$ of its potential schedule. Specifically under \textsf{Apt-Serve}, the potential schedule associated with a request $i$ in an iteration $e$ can be described as a tuple $(\alpha^{e}_i,\beta^{e}_i)$, where $\alpha^{e}_i$ is a binary variable indicating whether request $i$ is selected to composite the execution batch in the iteration $e$, and $\beta^{e}_i$ is also a binary variable indicating whether request $i$ is assigned with hidden cache usage.

Intuitively, the scheduling value $\mathrm{g}^{e}_i$ is quantified by assessing its contribution ($\alpha^{e}_{i}=1$) to reducing the sum of pending time across all requests handled by the system in iteration $e$ ($\forall i \in \mathrm{W}^e \cup \mathrm{R}^e$). Such a value is formally defined as follows:
\begin{align}
    \mathrm{g}^{e}_{i} &= \mathrm{p}^{e}_{i} - \beta^{e}_{i}(|\mathrm{W}^{e}|+|\mathrm{R}^{e}|)\mathrm{t}^{e}_{i}, \label{schedule_value} \\
     \mathrm{t}^{e}_{i} &= \mathrm{\rho} \mathrm{m}^{e}_{i}. \label{linear_model}
\end{align}
The reason for the first part $\mathrm{p}^{e}_{i}$ in Eq.\ref{schedule_value} is that if the request $i$ is scheduled at a target iteration $e$, its pending time in the next iteration $\mathrm{p}^{e+1}_i$ is refreshed with a relatively small number close to zero, thus contributing to reducing the sum of latency across all requests. While the second part $\beta^{e}_{i}(|\mathrm{W}^{e}|+|\mathrm{R}^{e}|)\mathrm{t}^{e}_{i}$ in Eq.\ref{schedule_value} represents a potential penalty if request $i$ is assigned with hidden cache for its schedule ($\beta^{e}_{i}=1$). Specifically, $\mathrm{t}^{e}_{i}$ is the extra linear transformation cost by its hidden cache usage. The reason for the scaling factor $|\mathrm{W}^{e}|+|\mathrm{R}^{e}|$ is that the extra cost of hidden cache usage of a single request actually causes a reduced batch execution speed (Section~\ref{sc3.1}), which further leads to an increase of latency perceivable by all requests ($\forall i \in \mathrm{W}^e \cup \mathrm{R}^e$). Furthermore, for the extra cost $\mathrm{t}^{e}_i$, it can be well approximated using a linear model. 
This approximation is reasonable as the time complexity of the linear transformation is proportional to the sequence length of request $i$. 
The coefficient $\rho$ in Eq.~\ref{linear_model} can be determined before activating the serving pipeline, involving a marginal preprocessing cost of approximately 30 seconds in practice. In this way, the request manager can estimate the extra cost by any amount of hidden cache usage during serving.

Additionally, the quantification model incorporates an SLO-aware fallback mechanism. For a given request $i$ in the inference iteration $e$, its tracked pending time $\mathrm{p}^{e}_{i}$ may exceed the latency SLOs. Including such a request in the running queue without proper consideration could block other requests that are still within their latency SLOs. This may lead to a chain reaction of SLO violations for numerous subsequent requests, as illustrated in Section~\ref{sc3.2}. To address this issue, when a request $i$ has already violated its SLOs in iteration $e$, the request manager substitutes its original scheduling value $\mathrm{g}^{e}_{i}$ with a near-zero constant for a priority demotion.

Using the quantification model, the request manager performs adaptive runtime scheduling (as detailed in Section~\ref{adaptive_scheduling}) to produce the final batch schedule 
\(\{(\alpha^{e}_i, \beta^{e}_i) \mid i \in \mathrm{W}^e \cup \mathrm{R}^e, \alpha^{e}_i = 1\}\) 
for the target inference iteration \(e\). This final schedule is then passed to the hybrid cache assigner, which, in coordination with the cache engine, oversees the low-level allocation of the corresponding cache for each scheduled request in the GPU memory during iteration \(e\).

\begin{figure}[t]
    \centering
    \includegraphics[width=0.5\columnwidth]{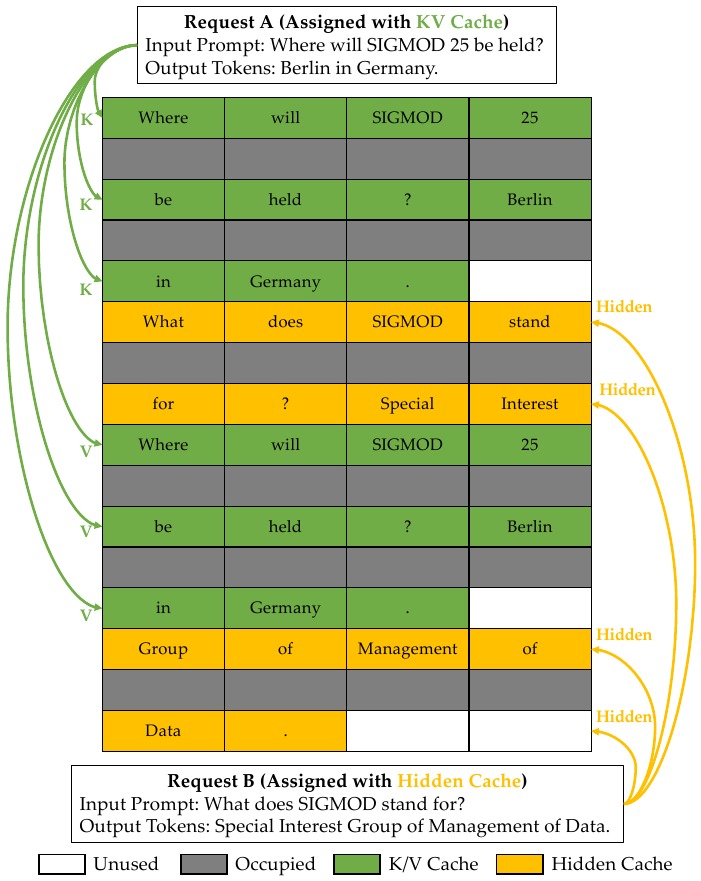}
    \caption{An illustration of managing both KV cache and hidden cache in the unified memory pool.}
    \label{fig:mem_pool}
\end{figure}
\subsection{Hybrid Cache Assigner and Cache Engine}\label{hybrid_design}

The hybrid cache assigner and cache engine manage a global memory pool for cache storage throughout the entire streaming serving process, following common practice in existing works~\cite{kwon2023efficient,agrawal2024taming,zhong2024distserve,Llumnix}. To ensure seamless hybrid cache utilization during serving, the hybrid cache assigner and cache engine closely coordinate to efficiently allocate different types of cache for various requests within the global memory pool.

\textbf{Workflow.} In each inference iteration, the hybrid cache assigner begins by receiving the finalized batch schedule from the request manager. Using this schedule, it creates a cache map for each request, detailing the physical location where the corresponding cache is stored in the global memory pool. These cache maps are then utilized by the cache engine to handle the storage and retrieval of caches during the inference process.

\textbf{Tailored Global Memory Pool.} To enable memory-efficient and flexible hybrid cache utilization during serving (Section~\ref{sc3.1}), we design a tailored global memory pool for the storage of both types of cache. First, since both KV cache and hidden cache can expand dynamically with the sequence length of target requests, organizing the global memory pool into a number of fixed-size blocks for storing both types of cache can help minimize memory fragmentation, thereby maximizing GPU memory utilization as illustrated in Section~\ref{serving_pre}. Second, to accommodate both types of cache within the global memory pool, a straightforward approach is to pre-allocate distinct storage spaces, taking into account their differing sizes. However, this naive storage strategy may lead to potential memory contention, limiting the flexibility to switch cache types when needed. For example, in a specific inference iteration, if a larger batch size is preferred, the system might encounter challenges due to insufficient storage space assigned for the hidden cache. Similarly, if a higher batch processing speed is desired, the system may be unable to meet this requirement because of inadequate space allocated for KV cache.

To address the problem, \textsf{Apt-Serve} \textit{jointly} manages the storage of KV cache and hidden cache over the \textit{unified} block-wise memory pool. Specifically, the subtlety lies in the granularity of cache blocks. Within the global memory pool by existing KV-cache-only systems, each cache block stores both the key $\mathbf{k^{\ell}_{(\cdot)}}$ and value vectors $\mathbf{v^{\ell}_{(\cdot)}}$ for a fixed number of tokens ($\ell=1,\dots,L$). While in \textsf{Apt-Serve}'s tailored global memory pool, each cache block stores either the key $\mathbf{k^{\ell}_{(\cdot)}}$, value $\mathbf{v^{\ell}_{(\cdot)}}$, or input $\mathbf{x^{\ell}_{(\cdot)}}$ vectors ($\forall \ell=1,2,...,L$) for a fixed number of tokens. This design leverages the fact that these vectors share the same dimension for each token position within a request. The unified memory pool thus enables the K cache, V cache, and hidden cache to occupy any cache block in a space-sharing manner, allowing a flexible cache type switch during serving.

\textbf{Cache Block Allocation.} Utilizing the finalized batch schedule $\{(\alpha^{e}_i, \beta^{e}_i)\mid i \in \mathrm{W}^e \cup \mathrm{R}^e, \alpha^{e}_i = 1\}$ from the request manager, the hybrid cache assigner first identifies the cache type ($\beta^{e}_i$) for a target scheduled request $i$ $(\alpha^{e}_i = 1)$, and generates or updates its cache map $\mathrm{c}^{e}_i$, depending on whether the request is in the prefill or decode iteration. The cache map $\mathrm{c}^{e}_i$ is a list that associates each token position of the cache for request $i$ with a specific cache block.

Given the cache maps from the hybrid cache assigner, the cache engine decides the actual cache allocation for the scheduled requests in the unified memory pool during the target inference iteration. Importantly, a single cache block stores cached vectors that are contiguous with respect to the token positions of a request. While the total blocks that contain the entire cache content for a request do not have to be contiguous within the unified memory pool.

Figure~\ref{fig:mem_pool} offers an example to illustrate how the KV cache and hidden cache for two different requests are jointly managed within the unified memory pool. In the example, the unified memory pool consists of 16 cache blocks (vertically) with block size 4 (horizontally), i.e., each block can accommodate K/V/hidden cache of 4 token positions. Request A in the figure is assigned with KV cache, which has 11 tokens (7 prompt tokens + 4 output tokens) in total. Its K cache occupies block 0, 2, and 4, while its V cache takes up block 8, 10, and 12 in the unified memory pool. By contrast, request B, with 14 tokens (6 prompts + 8 output) and assigned with hidden cache, places its hidden cache in block 5, 7, 13, and 15. In summary, the different types of cache for the two requests are divided into multiple cache blocks. The hybrid cache assigner selects the specific cache blocks for a target request on demand, by scanning the available unused cache blocks within the unified memory pool.

\section{Adaptive Runtime Scheduling in \textsf{Apt-Serve}}\label{adaptive_scheduling}
In this section, we elaborate on how the request manager in \textsf{Apt-Serve} adaptively derives the batch schedule in each inference iteration. In general, such an adaptive runtime scheduling process is composed of two stages. Firstly, the iteration type is decided, i.e., whether it is a prefill or decode iteration. Subsequently, the final batch schedule is determined, regarding not only which requests to composite the batch, but also which type of cache to assign to each scheduled request.

\textbf{Deciding the Iteration Type.} To decide the iteration type for a target inference iteration $e$, a common practice~\cite{kwon2023efficient} is to judiciously prioritize prefill iterations when sufficient memory is available to accommodate the cache of requests from the waiting queue $\mathrm{W}^e$. This allows for a larger batch size during decoding. However, this may cause significant generation stalls for requests already in the running queue $\mathrm{R}^e$, leading to latency violations (TBT) for those requests~\cite{agrawal2024taming}. To tackle this, the request manager adaptively determines the iteration type by assessing the real-time urgency of requests from both the waiting queue $\mathrm{W}^e$ and the running queue $\mathrm{R}^e$. Specifically, the iteration type is chosen based on which queue has a higher cumulative pending time, signaling greater urgency. Once the iteration type is decided, the candidate set of requests $\mathrm{U}^e$ ($\mathrm{U}^e =\mathrm{W}^e$ or $\mathrm{U}^e$) to derive the batch schedule is determined. 

\textbf{Deriving the Batch Schedule.}  Based on the potential schedules derived from the candidate set of requests $\mathrm{U}^e$, along with their corresponding quantified values (Section~\ref{request_manager}), the subsequent scheduling process at each inference iteration is formalized as a hybrid-cache-based scheduling problem (Definition~\ref{optimization_problem}). The goal is to maximize the reduction of overall pending time across all unfinished requests, subject to a memory constraint for cache storage. 

\begin{definition}[Hybrid-cache-based Scheduling Problem]\label{optimization_problem}
Given a set of candidate requests \( \mathrm{U}^{e} \) to be scheduled for execution, and a memory constraint \( \mathrm{M}^e \) for cache storage in a serving iteration $e$, the objective is to select a subset of schedules associated with candidate requests \( U^{e} \) that achieves the maximum sum of values:
\begin{align}
    \mathop{max} \quad & \sum_{i \in U^{e}} \mathrm{g}^{e}_i \alpha^{e}_i, \nonumber\\
    \text{s.t.} \quad & \sum_{i \in U^{e}} \left(1 - \frac{\beta^{e}_i}{2}\right) \mathrm{m}^{e}_i \alpha^{e}_i \leq \mathrm{M}^e, \label{constraint1}\\
    & \mathrm{g}^{e}_i = \mathrm{p}^{e}_i -  \beta^{e}_i (|\mathrm{W}^{e}|+|\mathrm{R}^{e}|) \mathrm{\rho} \mathrm{m}^{e}_i, \label{constraint2}\\
    & \alpha^{e}_i \in \{0,1\}, \beta^{e}_i \in \{0,1\}, \ \forall i \in \mathrm{U}^{e}. \label{constraint3}
\end{align}
\end{definition}

Note that \( \alpha^{e}_i \) and \( \beta^{e}_i \) are binary decision variables indicating whether request \( i \) is selected to composite the execution batch in iteration $e$, and whether it is assigned hidden cache usage. The pending time $\mathrm{p}^{e}_i$ and maximum memory requirement $\mathrm{m}^{e}_{i}$ for request $i$ at iteration $e$ are directly accessible, as they are tracked each iteration (Section~\ref{request_manager}). The scheduling value $\mathrm{g}^{e}_i$ for request $i$ is based on the pending time $\mathrm{p}^{e}_i$ (explained in Section~\ref{request_manager}) and can be calculated before scheduling. The memory constraint $\mathrm{M}^e$ in Eq.~\ref{constraint1} depends on the iteration type and the global memory pool size $\widetilde{\mathrm{M}}$. For prefill iterations, $\mathrm{M}^e = \widetilde{\mathrm{M}} - \sum_{i\in R^{e}} m^{e}_i$. Otherwise, $\mathrm{M}^e = \widetilde{\mathrm{M}}$. This memory constraint is computed in real-time during each iteration.

\textbf{Greedy-based Solution.} From Definition~\ref{optimization_problem}, it can be observed that the classic NP-hard 0-1 knapsack problem~\cite{martello1990knapsack} is a special case of the hybrid-cache-based scheduling problem (when \(\beta^{e}_i = 0\) for all \(i \in U^e\), which means no hidden cache usage is allowed). This confirms the inherent NP-hardness of the formulated optimization problem. To address such an NP-hard problem, the request manager in \textsf{Apt-Serve} uses a greedy-based approximate solution, which favors request schedules that have higher marginal gain of scheduling value per unit of memory consumption. For simplicity, we omit the superscript $e$ denoting the iteration in the following illustration.

For a given request \(i\), denote its current memory usage as \(\overline{\mathrm{m}}_{i}\), if \(\overline{\mathrm{m}}_{i}=0\), a marginal memory usage increase $\Delta \mathrm{m}_i$ to it involves assigning half of its maximum memory requirement \(\frac{\mathrm{m}_i}{2}\), indicating that request \(i\) is scheduled using hidden cache. Similarly, if \(\overline{\mathrm{m}}_{i} = \frac{\mathrm{m}_i}{2}\), the marginal memory usage increase $\Delta \mathrm{m}_i$ involves assigning the other half of its maximum memory requirement, indicating that request \(i\) is now scheduled using KV cache, given that it was previously scheduled with hidden cache. If \(\overline{\mathrm{m}}_{i} = \mathrm{m}_i\), assigning further memory does not provide any additional gain in value. Therefore, for a given request \(i\), its marginal gain in value $\mathrm{\theta}_i$ based on its current memory usage \(\overline{\mathrm{m}}_{i}\), is as follows:
\[
\mathrm{\theta}_i = 
\begin{cases}
    \frac{(\mathrm{p}_i - (|\mathrm{W}|+|\mathrm{R}|) \mathrm{\rho} \mathrm{m}_i)-0}{\mathrm{m}_i/2 - 0}=\frac{2\mathrm{p}_i}{\mathrm{m}_i} - 2(|\mathrm{W}|+|\mathrm{R}|)\mathrm{\rho}, \enskip &if \enskip \overline{\mathrm{m}}_i = 0;  \\
    \frac{\mathrm{p}_i - (\mathrm{p}_i - (|\mathrm{W}|+|\mathrm{R}|) \mathrm{\rho} \mathrm{m}_i)}{m_i - m_i/2}=2(|\mathrm{W}|+|\mathrm{R}|)\mathrm{\rho}, \enskip &if \enskip \overline{\mathrm{m}}_i = \frac{\mathrm{m}_i}{2}; \\
    0, \enskip &if \enskip \overline{\mathrm{m}}_i = \mathrm{m}_i.
\end{cases}
\]

Note that for a given request $i$ to be scheduled, there may be cases where its hidden cache usage imposes too much of a negative impact on other requests. This is evident when the marginal gain to the overall scheduling value from allocating just enough memory to hold its hidden cache (from $\overline{\mathrm{m}}_i=0$ to $\overline{\mathrm{m}}_i=\frac{\mathrm{m}_i}{2}$) is smaller than which by directly assigning the exact memory space to hold its KV cache (from $\overline{m}_i=0$ to $\overline{m}_i=m_i$). In such cases, its hidden cache usage is avoided, thus its marginal gain is refined as follows:
\[
\mathrm{\theta}_i = 
\begin{cases}
    \frac{\mathrm{p}_i-0}{\mathrm{m}_i - 0}=\frac{\mathrm{p}_i}{\mathrm{m}_i}, \enskip &if \enskip \overline{\mathrm{m}}_i = 0;  \\
    0, \enskip &if \enskip \overline{\mathrm{m}}_i = \mathrm{m}_i.
\end{cases}
\]

Assume the total number of candidate schedules (how much marginal memory usage increase to which requests) is $n$, all the possible marginal gain \(\mathrm{\theta}_1, \mathrm{\theta}_2, \dots, \mathrm{\theta}_n\) associated with each candidate request can be pre-recomputed. Then, we can derive the candidate schedule set $\Upsilon$ described as follow:
\begin{equation}
    \Upsilon=\{(\mathrm{\theta}_j, \mathrm{r}_j, \Delta \mathrm{m}_j, \mathrm{m}_{\mathrm{r}_j}) \mid j=1,2,...,n \}.\label{candidate_schedules}
\end{equation}
For the $j$-th candidate schedule in the set $\Upsilon$, $\theta_j$ represents its marginal gain in value. $\mathrm{r}_j$ corresponds to the request associated with the $j$-th candidate schedule. $\Delta \mathrm{m}_j$ indicates the marginal increase in memory usage required by this schedule, while $\mathrm{m}_{\mathrm{r}_j}$ denotes the maximum memory required by request $\mathrm{r}_j$ for the $j$-th candidate schedule. With the candidate schedule set $\Upsilon$ derived, together with the candidate request set $U$ and the memory constraint $M$, it is fed to a greedy-based scheduling process as input, to derive the final scheduling decisions $S$ as follow:
\begin{equation}
    S=\{(\alpha_i, \beta_i) \mid i \in U\}. \label{schedule_decisions}
\end{equation}

We further theoretically prove that such a solution has an approximation ratio of 2. Due to page limit, we move the details of the scheduling algorithm and relevant theoretical proof to online appendix\footnote{\url{https://github.com/eddiegaoo/Apt-Serve/blob/main/greedy_scheduling_appendix.pdf}}. Since the request manager in \textsf{Apt-Serve} dynamically forms request compositions based on scheduling outcomes that adapt to runtime information in each inference iteration, it is important to note that a request may need to switch cache types according to the scheduling result of a particular iteration. In such cases, \textsf{Apt-Serve} discards the existing cache and schedules a prefill iteration to recompute the cache in the required type.

\section{Experiments}
In this section, we first provide the implementation details of  \textsf{Apt-Serve} (Section~\ref{impl}) and the experiment setups (Section~\ref{exp_setup}). Next, we compare the effective throughput of \textsf{Apt-Serve} with three advanced inference serving systems (Section~\ref{main_exp}), and assess \textsf{Apt-Serve}'s robustness under varying request arrival patterns (Section~\ref{burstiness}). We also examine the effects of the hybrid cache and the adaptive runtime scheduling (Section~\ref{eval_hybrid}). We further evaluate the impact of \textsf{Apt-Serve}'s scheduling in depth (Section~\ref{schedule_analysis}), as well as the \textsf{Apt-Serve}'s generalization ability (Section~\ref{generalization}).  

\subsection{Implementation Details}\label{impl}
We implement \textsf{Apt-Serve}\footnote{Publicly available at: \url{https://github.com/eddiegaoo/Apt-Serve}} on top of the advanced open-source serving system vLLM~\cite{kwon2023efficient}, inheriting several \textit{de facto} systematic optimizations including FlashAttention~\cite{dao2022flashattention} and iteration-level batching~\cite{yu2022orca}. For the scheduler part, we implement runtime information checking and quantification and the adaptive scheduling algorithm to automatically decide the batch composition with the desired cache type for requests within the batch in each iteration, and the tailored block-wise memory pool of to support hybrid cache storage. For the inference engine, both the KV cache and hidden cache are stored in a block-wise format to allow flexible batch size configurations. However, irregular memory access can frequently occur due to the fragmented nature of the cache data, which is scattered across the physical memory space, even for a single request. Such fragmentation can increase inference latency by causing additional memory access overhead~\cite{miao2023towards}. To mitigate this, we devise a specialized CUDA kernel that optimizes block-wise hidden cache I/O operations, similar to the one used for KV cache~\cite{kwon2023efficient}. This kernel fuses reshaping with read/write operations, and enables efficient parallel access to fragmented cache on the GPU. For the model executor part, we use the NCCL~\cite{nccl} for tensor parallel communication among GPU workers as default in the original vLLM implementation.

\subsection{Experiment Setups}\label{exp_setup}
\textbf{Hardware Configurations \& Models.} We conduct all the experiments on a server with NVIDIA A100 GPUs, each with 40GB GPU memory, and the NVLink connections between GPUs are available. For the used models, following existing works~\cite{kwon2023efficient, zhong2024distserve}, we choose the OPT~\cite{zhang2022opt} model series that is a representative LLM family widely used in academia and industry. We use the default FP16 precision for each model, and the default tensor model parallelism~\cite{shoeybi2019megatron} when the model requires more than one GPU. Table~\ref{model_hardware} summarizes the model sizes and the corresponding hardware configurations.

\noindent\textbf{Datasets \& Workloads.}
To evaluate the effectiveness of \textsf{Apt-Serve}, we choose three typical LLM-based applications: chatbot, code-completion, and summarization with their respective benchmark datasets following prior works~\cite{kwon2023efficient,sharegpt}. For the chatbot application, we choose ShareGPT dataset~\cite{sharegpt}  consisting of a collection of user-shared conversations with ChatGPT~\cite{achiam2023gpt}. For the code-completion task, we choose HumanEval dataset~\cite{chen2021evaluating}, which features 164 programming problems, each accompanied by a function signature or docstring, designed to assess the performance of code completion models. For the summarization task, we choose LongBench dataset~\cite{bai2023longbench}, which consists of summarization of requests with longer prompts compared to the previous two datasets\footnote{We limit the sequence lengths in LongBench following prior works~\cite{kwon2023efficient, zhong2024distserve}, due to OPT’s absolute positional embedding only supporting a maximum length of 2048.}. 

\begin{figure}[t]
    \centering
    \includegraphics[width=\columnwidth]{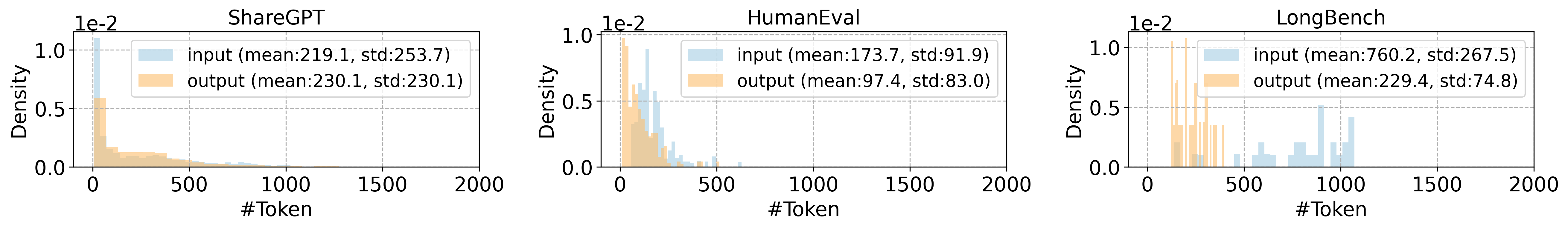}
    \caption{The input \& output length distributions of the sampled requests from ShareGPT, HumanEval and LongBench datasets.}\label{dataset_distribution}
\end{figure}

Following existing works~\cite{kwon2023efficient,zhong2024distserve,agrawal2024taming,Llumnix}, we create a distinct serving trace for each dataset respectively, by randomly sampling 1,000 requests from each dataset. We further generate corresponding request arrivals using Poisson distribution with different request rates, as these datasets do not include timestamps associated with the requests. Figure~\ref{dataset_distribution} shows the distribution of input length and output length of the sampled requests in each dataset. It can be noted that the distributions in different datasets vary a lot, as they correspond to different downstream tasks of real-time LLM inference serving.

\begin{table}[t]
    \centering
    \small 
    \caption{Model sizes and hardware configurations.}\label{model_hardware}
    \begin{tabular}{c|ccc}
         \toprule
         \textbf{Model Size} & \textbf{13B} & \textbf{30B} & \textbf{66B} \\
         \midrule
         \#GPUs & A100 & 2 $\times$ A100 & 4 $\times$ A100\\
          Total GPU Memory & 40GB & 80GB & 160GB\\
          Parameter Size & 26GB & 60GB & 132GB \\
          \bottomrule
    \end{tabular}
\end{table}

\begin{table}[t]
    \centering
    \small 
    \caption{TTFT \& P99 TBT SLOs (s) on different datasets and hardware configurations.}
    \label{slo_setting}
    \scalebox{0.9}{
    \begin{tabular}{c|cc|cc|cc}
        \toprule
         \textbf{Model Size} & \multicolumn{2}{c|}{\textbf{13B}} & \multicolumn{2}{c|}{\textbf{30B}} & \multicolumn{2}{c}{\textbf{66B}} \\
         SLOs & TTFT & P99 TBT & TTFT & P99 TBT & TTFT & P99 TBT \\
         \midrule
         ShareGPT & 1.0 & 1.0 & 1.5 & 1.0 & 2.0 & 1.0 \\
         HumanEval & 0.5 & 0.5 & 1 & 0.5 & 1.5 & 0.5 \\
         LongBench & 4.0 & 1.0 & 4.5 & 1.0 & 5.0 & 1.0\\
         \bottomrule
    \end{tabular}}
\end{table}

\noindent\textbf{Baselines.} We compare \textsf{Apt-Serve} to three representative state-of-the-art LLM inference serving systems.

{$\bullet$} \textbf{vLLM.} vLLM~\cite{kwon2023efficient} is an open-sourced LLM inference serving system that is extensively utilized in both academia and industry. It features iteration-level batching~\cite{yu2022orca} and employs the most advanced block-wise KV cache management, which significantly reduces memory fragmentation in KV cache allocation maximizing the attainable batch size under solely the KV cache usage. 

{$\bullet$} \textbf{Sarathi-Serve.} Sarathi-Serve~\cite{agrawal2024taming} is a recent state-of-the-art LLM inference serving system implemented atop vLLM, which is further equipped with a chunked prefill technique and an advanced iteration-level prefill-decode coalescing batching mechanism. It aims to improve the effective throughput by optimizing per-batch execution speed via better coordinating the computation resource utilization. The newly introduced designs by \textsf{Apt-Serve} in this paper are orthogonal to the ones in Sarathi-Serve, and can be combined together for a more enhanced effective throughput.

{$\bullet$} \textbf{DeepSpeed-FastGen.} DeepSpeed-FastGen~\cite{holmes2024deepspeed} is also a state-of-the-art LLM inference serving system equipped with the prefill-decode coalescing batching mechanism. Its optimizations are similar to the ones in Sarathi-Serve, but differs in the token composition strategy under the same token budget.

To ensure a fair comparison, we maintain consistent GPU memory utilization of cache storage for all the baselines and \textsf{Apt-Serve}.

\noindent\textbf{Metrics.} We focus on effective throughput. Following existing works~\cite{zhong2024distserve, agrawal2024taming}, we measure the SLO attainment rate (\%) using the same workloads with different request rates. The SLO attainment rate is calculated by the number of served user requests that satisfy both TTFT and P99 TBT SLOs divided by the total number of requests during the whole serving process. We can compare the effective throughput of different systems by checking the maximum request rate sustained under the same level of SLO attainment.

For different datasets and hardware configurations, we set the application-driven SLOs based on the same principle in existing work~\cite{zhong2024distserve}. Specifically, for chatbot and code-completion tasks, users usually demand not only immediate real-time response (i.e., the time to see the first token) but also satisfactory per-token generation speed. Therefore, both stringent TTFT and P99 TBT SLOs are required to be set for these two tasks. For the summarization task, as the input prompts are usually long requiring more overhead during the prefill phase, a relatively loose TTFT SLO is considered for this type of application. Besides, with the increase of model size, the model execution latency also increases. Therefore, for larger models, both TTFT SLO and P99 TBT SLOs are slightly relaxed compared to the models of the smaller sizes. Table~\ref{slo_setting} presents the detailed SLOs for different applications under different hardware settings used in the main experiments (Section~\ref{main_exp}). 

\begin{figure}[t]
    \centering
    \begin{subfigure}[b]{\columnwidth}
        \centering
        \includegraphics[width=\textwidth]{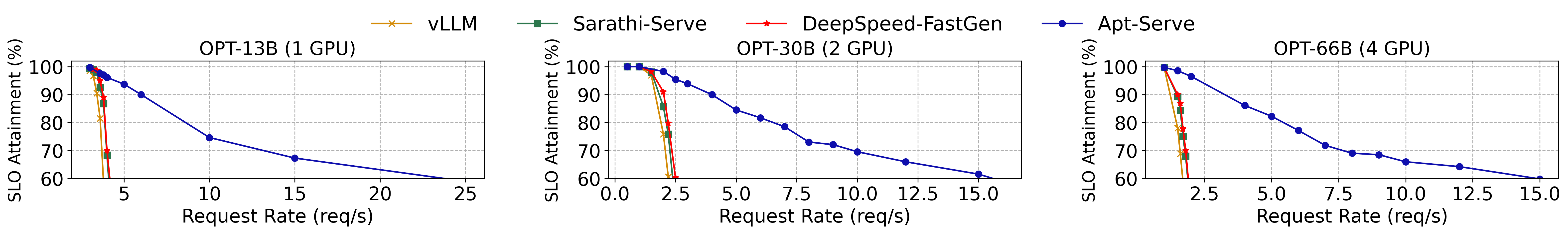}
        \caption{ShareGPT}\label{sharegpt_throughput}
    \end{subfigure}
    
    \begin{subfigure}[b]{\columnwidth}
        \centering
        \includegraphics[width=\textwidth]{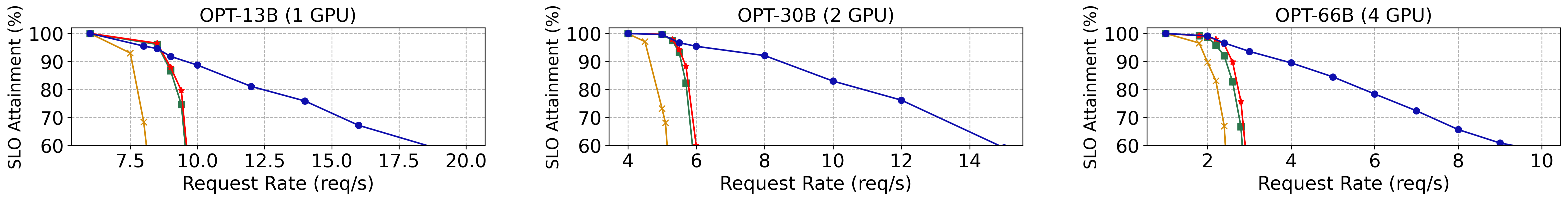}
        \caption{HumanEval}\label{humaneval_throughput}
    \end{subfigure}
    \begin{subfigure}[b]{\columnwidth}
        \centering
        \includegraphics[width=\textwidth]{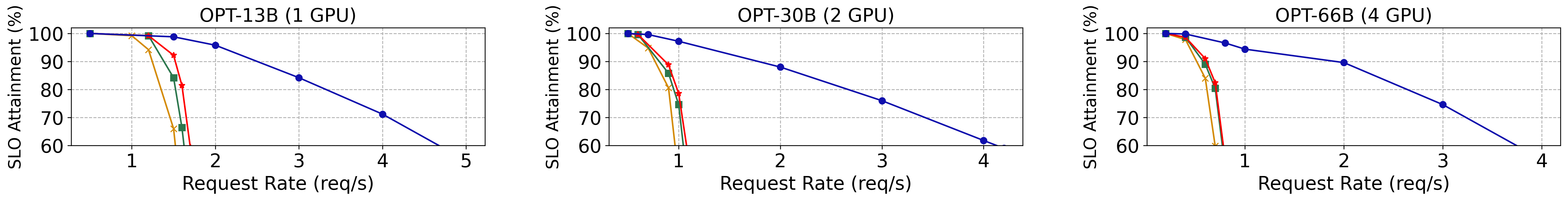}
        \caption{LongBench}\label{longbench_throughput}
    \end{subfigure}
    \caption{Effective throughput comparison among vLLM, Sarathi-Serve, DeepSpeed-FastGen and \textsf{Apt-Serve} on ShareGPT, HumanEval, and LongBench datasets with different models.}
    \label{fig:throughput}
\end{figure}
\subsection{Main Results: Effective Throughput} \label{main_exp}
We compare the effective throughput between all three baselines and \textsf{Apt-Serve} on three models of different sizes, and three datasets of distinct distributions, under different hardware configurations. The results are summarized in Figure~\ref{fig:throughput}. Within each subfigure, the curves illustrate that as the request rate rises, an increasing number of requests fail to meet the latency requirements, resulting in a decrease in SLO attainment for all systems. Notably, the baseline systems are highly sensitive to even a moderate increase in request rate beyond a certain threshold. In contrast, \textsf{Apt-Serve} remains resilient under high request rates, avoiding a sudden collapse in SLO attainment thanks to its hybrid cache and adaptive scheduling designs. Compared to vLLM, Sarathi-Serve, and DeepSpeed-FastGen, \textsf{Apt-Serve} achieves 2.3$\times$, 1.9$\times$, and 1.8$\times$ higher average request rates respectively, with peak values reaching 4.0$\times$, 3.4$\times$, and 3.3$\times$ at 90\% SLO attainment. At 60\% SLO attainment, \textsf{Apt-Serve} can handle 4.9$\times$, 4.4$\times$, and 4.3$\times$ higher request rates on average, with maximum values up to 8.8$\times$, 7.9$\times$, and 7.5$\times$, respectively. In the following, we provide a detailed comparative analysis across different datasets, exploring how \textsf{Apt-Serve}'s optimizations perform with different request workloads.

On ShareGPT, \textsf{Apt-Serve} demonstrates a significant improvement in effective throughput.  Figure~\ref{sharegpt_throughput}, at the 90\% SLO attainment threshold, \textsf{Apt-Serve} can handle 2.3$\times$, 2.0$\times$, and 1.9$\times$ higher request rates on average compared to vLLM, Sarathi-Serve, and DeepSpeed-FastGen. At the 60\% SLO attainment threshold, this advantage increases, with \textsf{Apt-Serve} sustaining 7.4$\times$, 6.8$\times$, and 6.4$\times$ higher request rates on average. As seen in Figure~\ref{dataset_distribution}, ShareGPT has the highest mean output lengths, resulting in longer cache lifetimes for ongoing requests, which further makes incoming request wait longer for available cache space as the request rate rises. \textsf{Apt-Serve}'s hybrid cache design effectively accommodates more requests within the same memory budget, reducing TTFT SLO violations. Additionally, the high variance in both input and output lengths in the ShareGPT dataset makes \textsf{Apt-Serve}'s adaptive scheduling particularly beneficial. Since it can dynamically adjust the request composition in each inference iteration, reducing unnecessary SLO violations under the same memory budget.

On HumanEval, \textsf{Apt-Serve} shows a moderate improvement in effective throughput as seen in Figure~\ref{humaneval_throughput}. On average, \textsf{Apt-Serve} handles 1.7$\times$, 1.4$\times$, and 1.4$\times$ higher average request rates at the 90\% SLO threshold, and 3.0$\times$, 2.6$\times$, and 2.5$\times$ higher at the 60\% threshold, compared to vLLM, Sarathi-Serve, and DeepSpeed-FastGen. The performance difference is expected, as HumanEval has smaller average output lengths and lower variance in both input and output lengths (Figure~\ref{dataset_distribution}). These characteristics reduce the impact of \textsf{Apt-Serve}'s optimizations, as per-request computational and memory demands are lower. We also observe that Sarathi-Serve and FastGen perform better on HumanEval, likely due to the short per-request cache lifetime (due to smaller output lengths) and their prefill-decode coalescing batching, which reduces the generation stall for decode requests and frees up cache space faster. However, when per-request cache lifetime increases (e.g., on ShareGPT), this optimization alone is inadequate to boost effective throughput.

On LongBench, \textsf{Apt-Serve} also shows a significant boost in effective throughput. As indicated in Figure~\ref{longbench_throughput}, on average, \textsf{Apt-Serve} can sustain 2.8$\times$, 2.5$\times$, and 2.3$\times$ higher request rates at the 90\% SLO attainment threshold, and 4.2$\times$, 3.9$\times$, and 3.8$\times$ higher request rates at the 60\% threshold compared to the three baselines respectively. LongBench shares similar characteristics with ShareGPT in terms of large mean output lengths, but it also has significantly larger mean input lengths, indicating higher per-request cache memory consumption. In such cases, \textsf{Apt-Serve}'s hybrid cache design lowers per-request memory consumption, reducing the number of pending requests. Additionally, the high variance in input lengths on the LongBench dataset makes \textsf{Apt-Serve}'s adaptive scheduling highly effective, similar to the benefit observed on ShareGPT.

\subsection{Evaluation on Robustness} \label{burstiness}
\begin{figure*}[t]
    \centering
    \includegraphics[width=\columnwidth]{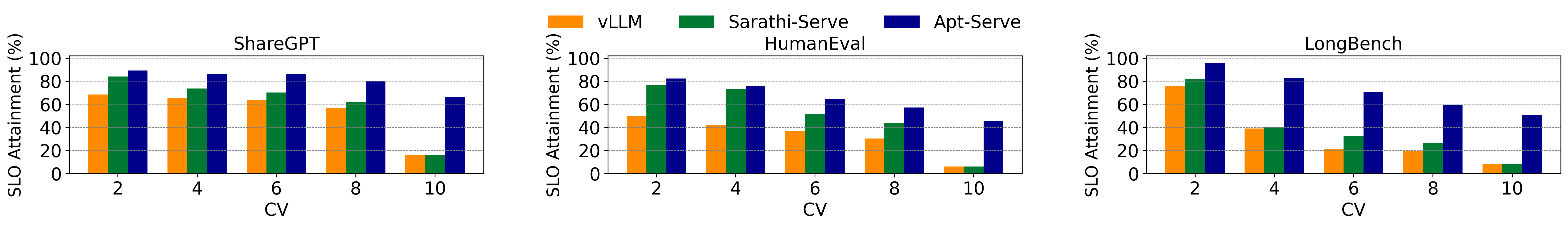}
    \caption{The SLO attainment (\%) comparison among vLLM, Sarathi-Serve and \textsf{Apt-Serve} under different levels of burstiness on ShareGPT, HumanEval, and LongBench datasets.}
    \label{fig:burstiness}
\end{figure*}
We evaluate the robustness of \textsf{Apt-Serve} and the baselines on distinct request arrival patterns of the same request rate.
We use Gamma distribution to simulate request arrivals across three different datasets, keeping the average arrival rate fixed while varying the burstiness of the arrivals, which is controlled by the coefficient of variation (CV) in the Gamma distribution. A higher CV indicates more bursty arrivals. We choose OPT-13B as a representative model, as the results for other models follow a similar trend. The request rates are set to 3.8 req/s for ShareGPT, 9.0 req/s for HumanEval, and 1.5 req/s for LongBench, where the original performance of \textsf{Apt-Serve} is close to that of the strongest baseline Sarathi-Serve in the main experiments. From Figure~\ref{fig:burstiness}, we observe that on all datasets, the SLO attainment for all systems declines as request burstiness increases. However, \textsf{Apt-Serve} consistently outperforms the baselines against higher burstiness, with the performance gap widening as burstiness increases. Overall, \textsf{Apt-Serve} achieves up to 7.5$\times$ higher SLO attainment under bursty request conditions compared to the baselines.

\begin{table}[t]
    \centering
    \small 
    \caption{The SLO attainment (\%) of \textsf{Apt-Serve} with KV cache and with hybrid cache under differed request rates and arrival burstiness (CV) on ShareGPT and LongBench datasets.}\label{hybrid_cache_effect}
    \begin{tabular}{c|c|c|cc}
         \toprule
         \textbf{Dataset} & \textbf{Request Rate} & \textbf{CV} & \textbf{KV Cache} & \textbf{Hybrid Cache} \\
         \midrule
         \multirow{6}{*}{ShareGPT} & \multirow{3}{*}{3} & 1 & 99.5 & 99.5 \\
         {} & {} & 5 & 94.7 & 95.9 \\
         {} & {} & 10 & 75.1 & 78.2 \\
         \cline{2-5}
         {} & \multirow{3}{*}{6} & 1 & 65.7 & 67.3 \\
         {} & {} & 5 & 66.7 & 67.7 \\
         {} & {} & 10 & 57.1 & 58.4 \\
         \midrule
         \multirow{6}{*}{LongBench} & \multirow{3}{*}{1.5} & 1 & 96.6 & 98.2 \\
         {} & {} & 5 & 70.0 & 76.0 \\
         {} & {} & 10 & 43.4 & 50.8 \\
         \cline{2-5}
         {} & \multirow{3}{*}{3} & 1 & 70.0 & 77.6 \\
         {} & {} & 5 & 58.8 & 64.8 \\
         {} & {} & 10 & 35.6 & 42.5 \\
         \bottomrule
    \end{tabular}
\end{table}
\begin{table}[t]
    \centering
    \small 
    \caption{The SLO attainment (\%) of \textsf{Apt-Serve} with FCFS and with adaptive scheduling under differed request rates and arrival burstiness (CV) on ShareGPT and LongBench.}\label{schedule_policy_effect}
    \begin{tabular}{c|c|c|cc}
         \toprule
         \textbf{Dataset} & \textbf{Request Rate} & \textbf{CV} & \textbf{FCFS} & \textbf{Adaptive} \\
         \midrule
         \multirow{6}{*}{ShareGPT} & \multirow{3}{*}{3} & 1 & 26.4 & 99.5 \\
         {} & {} & 5 & 59.7 & 95.9 \\
         {} & {} & 10 & 11.9 & 78.2 \\
         \cline{2-5}
         {} & \multirow{3}{*}{6} & 1 & 20.0 & 67.3 \\
         {} & {} & 5 & 18.1 & 67.7 \\
         {} & {} & 10 & 9.9 & 58.4 \\
         \midrule
         \multirow{6}{*}{LongBench} & \multirow{3}{*}{1.5} & 1 & 11.8 & 98.2 \\
         {} & {} & 5 & 10.8 & 76.0 \\
         {} & {} & 10 & 5.4 & 50.8 \\
         \cline{2-5}
         {} & \multirow{3}{*}{3} & 1 & 4.8 & 77.6 \\
         {} & {} & 5 & 4.6 & 64.8 \\
         {} & {} & 10 & 4.2 & 42.5 \\
         \bottomrule
    \end{tabular}
\end{table}

\subsection{Ablation Study}\label{eval_hybrid}
We conduct case studies to evaluate the impact of the hybrid cache design in \textsf{Apt-Serve} using the OPT-13B model on the ShareGPT and LongBench datasets. We use Gamma distribution to simulate request arrivals and vary both request rates and CVs. Specifically, we compare the performance of \textsf{Apt-Serve} with and without the hybrid cache (using only the KV cache) while retaining the adaptive scheduling design. The results are presented in Table~\ref{hybrid_cache_effect}. It is clear that \textsf{Apt-Serve} consistently achieves higher SLO attainment when utilizing the hybrid cache. Such performance gain becomes more prominent with higher request rate, burstier request load and longer requests, as the hybrid cache enables \textsf{Apt-Serve} to enlarge batch size more flexibly during the serving process. 

We also analyze the effect of scheduling policy in \textsf{Apt-Serve} using the same experimental setup described above. We compare \textsf{Apt-Serve}'s performance with its original adaptive scheduling policy, described in Section~\ref{adaptive_scheduling}, against the FCFS policy commonly used in other systems. As shown in Table~\ref{schedule_policy_effect}, the FCFS policy significantly degrades \textsf{Apt-Serve}'s performance, leading to poor SLO attainment. This is because the FCFS policy enforces rigid scheduling outcome, while \textsf{Apt-Serve}'s scheduling policy intelligently leverages runtime information to make adaptive scheduling decisions.

\subsection{Analysis of the Apt-Serve's Scheduling}\label{schedule_analysis}
\begin{figure}[t]
    \centering
    \begin{subfigure}[b]{\columnwidth}
        \centering
        \includegraphics[width=\textwidth]{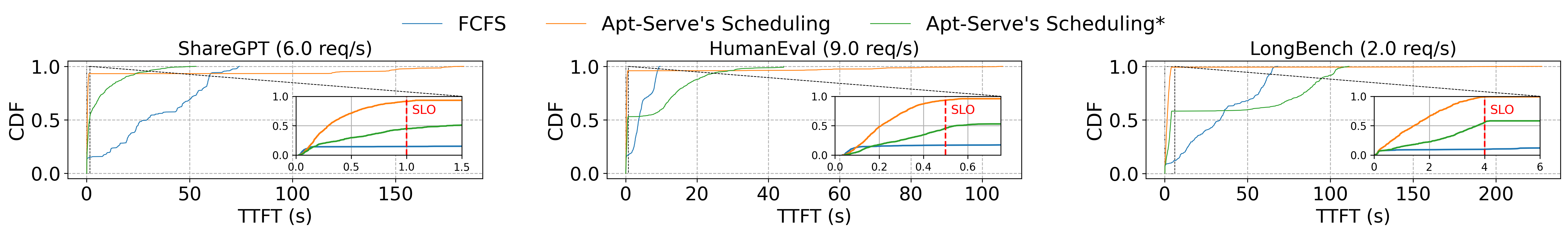}
        \caption{TTFT CDF}\label{ttft_cdf}
    \end{subfigure}
    
    \begin{subfigure}[b]{\columnwidth}
        \centering
        \includegraphics[width=\textwidth]{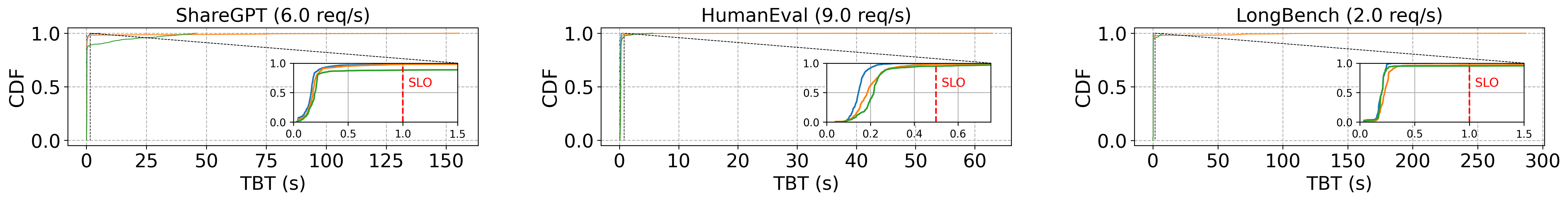}
        \caption{TBT CDF}\label{tbt_cdf}
    \end{subfigure}
    \caption{Request TTFT and P99 TBT distributions on ShareGPT (6.0 req/s), HumanEval (9.0 req/s), and LongBench (2.0 req/s) by FCFS scheduling, \textsf{Apt-Serve}'s scheduling and \textsf{Apt-Serve}'s scheduling*.}
    \label{fig:cdfs}
\end{figure}
We further analyze \textsf{Apt-Serve}'s scheduling, focusing on its impact on latency percentiles and its scalability.

\textbf{Effect on the Request Latency Distributions}. We analyze the cumulative distribution functions (CDFs) of TTFT and TBT for both FCFS scheduling and \textsf{Apt-Serve}'s scheduling, as illustrated in Figure~\ref{fig:cdfs}. The comparison is conducted at request rates of 6.0 req/s, 9.0 req/s, and 2.0 req/s on ShareGPT, HumanEval, and LongBench using the OPT-13B model. At these rates, \textsf{Apt-Serve}'s scheduling demonstrates its ability to ensure that the majority of requests meet their SLO criteria, achieving over 90\% SLO attainment. In contrast, FCFS scheduling results in a severe performance degradation, with less than 30\% SLO attainment. While \textsf{Apt-Serve} offers superior performance, it may cause a small fraction of requests (10\%) experience starvation, as shown by high tail latency. This occurs due to the SLO-aware fallback mechanism (described in Section~\ref{request_manager}), where requests exceeding their latency SLOs have their scheduling values aggressively reduced to near-zero for priority demotion, aiming at maximizing SLO attainment. Despite this, \textsf{Apt-Serve} remains adaptable, allowing further tradeoffs between SLO attainment and tail latency. For example, a feasible way is to apply the decaying factor to the scheduling values of SLO-violated requests, instead of directly reducing them to near-zero. We present an exemplar result (\textsf{Apt-Serve}'s Scheduling*) with a uniform decaying factor of 0.4 across all datasets in Figure~\ref{fig:cdfs}. We can observe that such an alternative configuration can achieve a significant reduction in tail latency compared to the original \textsf{Apt-Serve}'s scheduling, while still outperforming FCFS in terms of SLO attainment.
\begin{table}[t]
    \centering
    \small 
    \caption{The execution time (ms) of \textsf{Apt-Serve}'s scheduling algorithm against the number of candidate requests.}\label{schedule_scalability}
    \begin{tabular}{c|cccccc}
         \toprule
         \textbf{\# Request} & 50 & 100 & 200 & 400 & 800 & 1600 \\
         \midrule
         \textbf{Time (ms)} & 0.3 & 0.5 & 1.0 & 2.1 & 4.8 & 10.8
         \\
         \bottomrule
    \end{tabular}
\end{table}

\textbf{Scalability in Terms of Candidate Requests.} We also evaluate the execution time of \textsf{Apt-Serve}'s scheduling algorithm as the number of candidate requests increases using the OPT-13B model on a single GPU. The results, summarized in Table~\ref{schedule_scalability}, reveal that the algorithm is computationally efficient. Even when scheduling up to 1.6K requests, the incurred overhead is minimal—only 10.8 milliseconds—compared to the practical computation time. For example, a single decode iteration with 50 requests using the OPT-13B model takes approximately 120 milliseconds.

\subsection{Generalization Study}\label{generalization}

We further investigate \textsf{Apt-Serve}'s generalization capabilities, focusing on (1) its ability to integrate with other optimization techniques, and (2) its performance in ultra-long context scenarios.

\begin{figure}[t]
    \centering
    \includegraphics[width=\columnwidth]{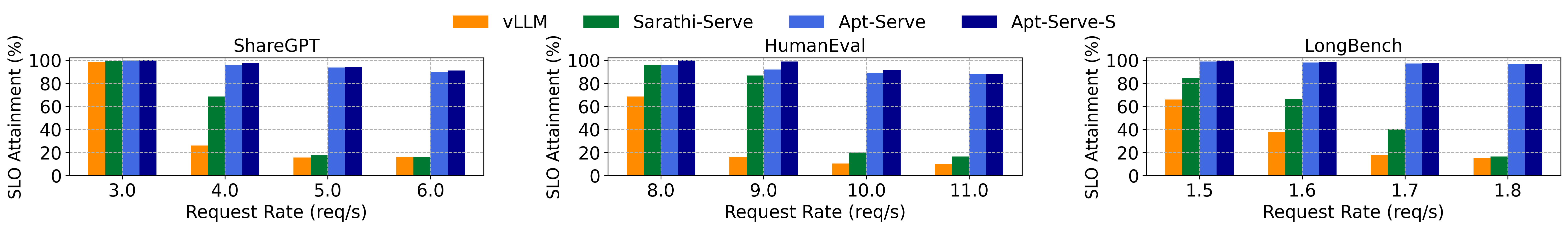}
    \caption{The SLO attainment (\%) comparison among vLLM, Sarathi-Serve, \textsf{Apt-Serve} and \textsf{Apt-Serve-S} under different request rates on ShareGPT, HumanEval, and LongBench datasets.}
    \label{fig:generalization_res}
\end{figure}

\begin{table}[t]
    \centering
    \small 
    \caption{The input \& output lengths statistics of three ultra-long datasets, WikiText, Arxiv, and BookCorpus.}\label{longcontext_stat}
    \begin{tabular}{c|ccc|ccc}
         \toprule
         \multirow{2}{*}{\textbf{Dataset}} & \multicolumn{3}{c|}{\textbf{Input Length}} & \multicolumn{3}{c}{\textbf{Output Length}} \\
         \cline{2-7}
         {} & Max & Median & Mean & Max & Median & Mean 
         \\
         \midrule
         WikiText & 1840 & 871 & 914 & 992 & 552 & 521 \\
         Arxiv & 19600 & 6853 & 7812 & 9754 & 226 & 420  \\
         BookCorpus & 23706 & 14781 & 16944 & 299 & 221 & 185 \\
         \bottomrule
    \end{tabular}
\end{table}
\textbf{Generalization with Other Techniques.} As \textsf{Apt-Serve}'s optimizations were initially built on vLLM, we further extend them on top of the Sarathi-Serve's optimizations, such as chunked prefill and coalesced batching of prefill and decode requests. This removes the need for the iteration type decision in Section~\ref{adaptive_scheduling}, focusing the scheduling process on request composition (\textsf{Apt-Serve-S}). We further compare vLLM, Sarathi-Serve, \textsf{Apt-Serve}, and \textsf{Apt-Serve-S} using the OPT-13B model across ShareGPT, HumanEval, and LongBench datasets (SLOs in Table~\ref{slo_setting}). The results in Figure~\ref{fig:generalization_res} show that \textsf{Apt-Serve-S} not only outperforms the baseline Sarathi-Serve, but also the original \textsf{Apt-Serve} (only integrating the vLLM's optimizations). Such a phenomenon is reasonable, as integrating Sarathi-Serve's optimizations enables better computation resource utilization. These findings suggest \textsf{Apt-Serve}'s techniques could be applicable to other emerging methods~\cite{zhong2024distserve,oh2024exegpt,patel2024splitwise,Llumnix}, which we leave for future work.

\begin{figure}[t]
    \centering
    \begin{subfigure}[b]{\columnwidth}
        \centering
        \includegraphics[width=\textwidth]{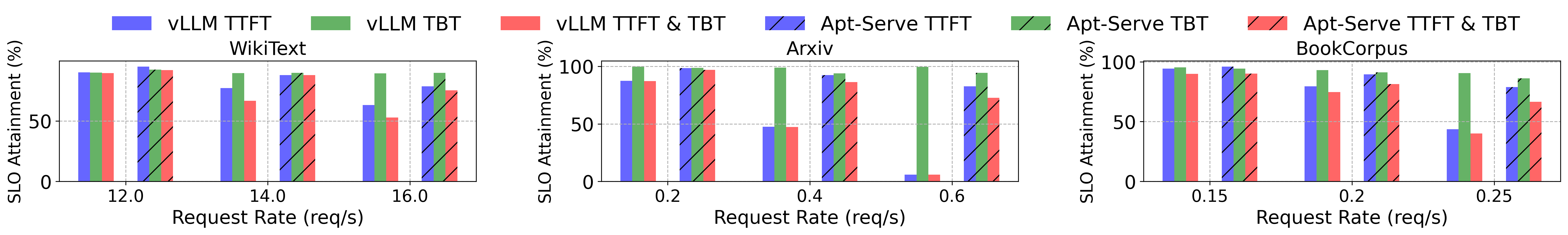}
        \caption{LLaMA3-8B-Instruct262K}\label{llama_throughput}
    \end{subfigure}
    \begin{subfigure}[b]{\columnwidth}
        \centering
        \includegraphics[width=\textwidth]{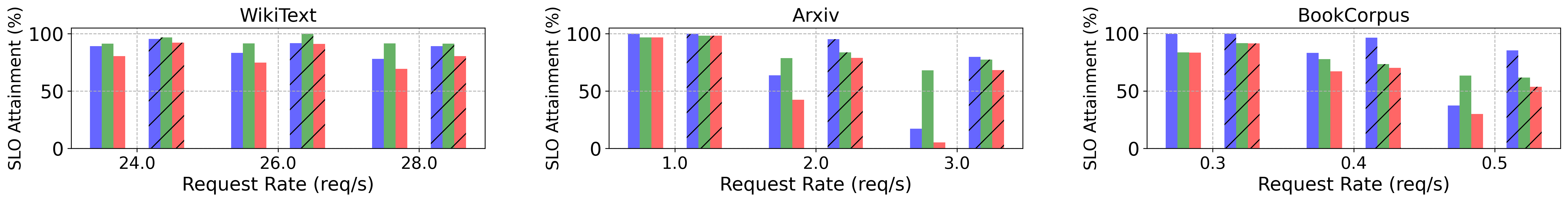}
        \caption{Yi-6B-200K}\label{yi_throughput}
    \end{subfigure}
    \caption{The SLO attainment (\%) comparison between vLLM and \textsf{Apt-Serve} on WikiText, Arxiv, and BookCorpus datasets with two different models.}
    \label{fig:longcontext}
\end{figure}
\textbf{Generalization to Ultra-Long Context.} Due to the 2048-token context limitation of the OPT model family, we evaluate \textsf{Apt-Serve} and vLLM using two models with extended context capabilities: LLaMA3-8B-Instruct262K and Yi-6B-200K. We test on three ultra-long context datasets—WikiText, Arxiv, and BookCorpus—sampling serving traces similar to Section~\ref{main_exp}. The input and output statistics for the sampled requests are shown in Table~\ref{longcontext_stat}. Experiments were run on 1 GPU, 2 GPUs, and 4 GPUs for the datasets, respectively, to handle larger context lengths and ensure batching of at least 10 requests per iteration. We set a relaxed TTFT SLO of 10 seconds in response to ultra-long prompts by these datasets, while maintaining the strict 1 second P99 TBT SLO.

The results in Figure~\ref{fig:longcontext} show that \textsf{Apt-Serve} outperforms vLLM in SLO attainment, especially for TTFT, which is the focus of its optimizations. It is noteworthy that maintaining TBT SLO in the ultra-long context scenarios is rather challenging. For example, on BookCorpus with Yi-6B-200K at 0.5 req/s, both systems struggle to exceed 60\% TBT SLO attainment. This is due to interference between prefill and decode iterations, which is worsened by long prompts in ultra-long context scenarios~\cite{zhong2024distserve}. Integrating disaggregated distributed architectures~\cite{zhong2024distserve,oh2024exegpt,patel2024splitwise} could help address this and offers a promising avenue for future research.

\section{Related Work}
\textbf{LLM Inference Optimizations.} The overlap between data management and machine learning has attracted increasing attention, leading to a surge of system related studies from the data management community~\cite{gao2024simple,gao2024etc, zhang2023ducati,zhang2022feature,zhao2025memo,li2024daha,miao2023sdpipe, miao2023galvatron, peng2022sancus, miao2021het, miao2022hetgmp,miao2021heterogeneity,isenko2022my,li2023orca,li2023zebra,mohan2021analyzing, um2023fastflow, wang2024improving,xiang2025capsule,zhou2023narrow,li2023early}. Specifically, Inference related system optimizations~\cite{lu2018accelerating,yangoptimizing2022,gao2024accelerating,yang2022demonstration,liu2024pp,zhang2023inferturbo,zhou2021accelerating,Biathlon24,Kang20,Sirin24,fang2020optimizing,fang2024stile,fang2021eto} are critical to the efficient deployment of AI-based services. For LLM inference, several advanced techniques have been proposed to improve inference performance. 
FlashAttention~\cite{dao2022flashattention} is a widely adopted method that leverages tiling and kernel optimizations to reduce I/O costs, significantly improving the speed of attention computation during inference. Furthermore, model compression and quantization techniques~\cite{liu2023deja,xia2023flash, xia2024quant,song2023powerinfer} have been employed to lower inference latency, albeit often at the cost of reduced model accuracy. To enhance efficiency during the decode phase, the use of KV caching~\cite{pope2023efficiently} has become a standard practice. However, addressing the substantial memory consumption associated with the KV cache has led to research on KV cache compression and quantization~\cite{zhang2024pqcache,zhang2024h2o,liu2024scissorhands,liu2024cachegen}. While these methods reduce memory usage, they also introduce information loss, resulting in performance degradation similar to that seen with model compression and quantization techniques. 
In contrast, the hidden cache introduced in \textsf{Apt-Serve} maintains model effectiveness by preserving comprehensive historical information through additional computation.

\noindent\textbf{LLM Online Serving Optimizations.} Optimizing online serving of LLMs is crucial for maintaining scalability, ensuring high throughput, and adhering to latency service level objectives (SLOs) in response to streaming request traffic. Orca~\cite{yu2022orca} introduces iteration-level batching, where requests can dynamically join or exit a batch at each iteration, rather than relying on traditional run-to-completion batching, thereby improving GPU utilization by enabling larger batch sizes. vLLM~\cite{kwon2023efficient} tackles batch size optimization by implementing block-wise KV cache management, reducing cache fragmentation. Llumnix~\cite{Llumnix} optimizes memory utilization across multiple instances by employing dynamic KV cache migration, allowing for larger batch processing. Sarathi-Serve~\cite{agrawal2024taming} and FastGen~\cite{holmes2024deepspeed} increase throughput by adopting chunked-prefill and prefill-decode coalescing batching techniques to maximize computational resource usage. Disaggregated hardware solutions such as DistServe~\cite{zhong2024distserve}, SplitWise~\cite{patel2024splitwise}, and ExeGPT~\cite{oh2024exegpt} further enhance throughput by distributing prefill and decode tasks across separate GPUs. Additionally, FastServe~\cite{wu2023fast} reduces request completion times through a preemptive time-slicing mechanism, while SpotServe~\cite{miao2024spotserve} leverages dynamic reparallelization to provide cost-efficient serving on preemptible cloud instances. The strategies employed by \textsf{Apt-Serve} complement these existing approaches and can be integrated with them to further improve online serving performance. Some other works further propose learning-based predictions for output information to assist scheduling~\cite{fu2024efficient,jin2023s}. $S^3$~\cite{jin2023s} predicts exact output lengths, while Fu \textit{et al.}~\cite{fu2024efficient} predicts the relative rank of output lengths. These prediction-based methods can be integrated with \textsf{Apt-Serve} to enable interval-level scheduling decisions, potentially improving scheduling outcomes. Since this requires a new formulation of the scheduling problem, we leave it as future work.

\section{Conclusion}
This paper presents \textsf{Apt-Serve}, a scalable framework aimed at improving effective throughput in LLM inference serving. We identify two major factors that limit effective throughput: the extensive use of KV cache and the First-Come-First-Serve Request Scheduling policy, both of which lead to a sharp decline in TTFT SLO attainment as request rate increases. To tackle the bottlenecks, \textsf{Apt-Serve} introduces a novel hybrid cache scheme the combines the advantages of the computation-efficient KV cache and the memory-efficient hidden cache, enabling larger batch sizes and reducing delays for incoming requests. Furthermore, \textsf{Apt-Serve} devises an efficient runtime scheduling mechanism that dynamically optimizes the timing and cache allocation for each request. Extensive experiments on multiple datasets and LLMs demonstrate that \textsf{Apt-Serve} can boost effective throughput by up to 8.8$\times$ compared to state-of-the-art LLM inference serving systems. For the future work, we plan to generalize Apt-Serve's designs to the multi-instance scenario, incorporating a more comprehensive multi-dimensional scheduling problem formulation. 

\begin{acks}
The authors would like to thank the anonymous reviewers for their insightful reviews. Lei Chen’s work is partially supported by National Key Research and Development Program of China Grant No. 2023YFF0725100, National Science Foundation of China (NSFC) under Grant No. U22B2060, Guangdong-Hong Kong Technology Innovation Joint Funding Scheme Project No. 2024A0505040012, the Hong Kong RGC GRF Project 16213620, RIF Project R6020-19, AOE Project AoE/E-603/18, Theme-based project TRS T41-603/20R, CRF Project C2004-21G, Guangdong Province Science and Technology Plan Project 2023A0505030011, Guangzhou municipality big data intelligence key lab, 2023A03J0012, Hong Kong ITC ITF grants MHX/078/21 and PRP/004/22FX, Zhujiang scholar program 2021JC02X170, Microsoft Research Asia Collaborative Research Grant, HKUST-Webank joint research lab and 2023 HKUST Shenzhen-Hong Kong Collaborative Innovation Institute Green Sustainability Special Fund, from Shui On Xintiandi and the InnoSpace GBA. Yanyan Shen’s work is supported by the National Key Research and Development Program of China (2022YFE0200500), Shanghai Municipal Science and Technology Major Project (2021SHZDZX0102), the Tencent Wechat Rhino-Bird Focused Research Program, and SJTU Global Strategic Partnership Fund (2021 SJTU-HKUST). 
\end{acks}


\bibliographystyle{ACM-Reference-Format}
\bibliography{citation}

\end{document}